\documentclass[10pt,twocolumn,letterpaper]{article}

\usepackage[pagenumbers]{cvpr} 

\usepackage{url}            
\usepackage{booktabs}       
\usepackage{amsfonts}       
\usepackage{nicefrac}       
\usepackage{microtype}      
\usepackage{xcolor}         
\usepackage{tabularx}
\usepackage{subcaption}
\usepackage{multicol}
\usepackage{multirow}
\usepackage{tabularray}
\usepackage{amsmath}
\usepackage{amssymb}
\usepackage{siunitx}
\usepackage{colortbl}
\usepackage{graphicx}
\usepackage{lipsum}
\usepackage{afterpage}
\usepackage{placeins}
\usepackage{pgfplots,gincltex}
\usepgfplotslibrary{groupplots}
\pgfplotsset{compat=1.6}
\usepackage{tikz}
\usetikzlibrary{arrows.meta}
\usetikzlibrary{intersections}   
\usepgfplotslibrary{fillbetween} 
\usepackage[accsupp]{axessibility}  

\usepackage{balance}

\definecolor{cvprblue}{rgb}{0.21,0.49,0.74}
\definecolor{pltblue}{RGB}{174, 199, 232}
\definecolor{pltorange}{RGB}{255, 229, 204}
\definecolor{pltgreen}{RGB}{204, 229, 204}
\definecolor{pltred}{RGB}{229, 204, 204}
\definecolor{pltpurple}{RGB}{239, 218, 230}

\definecolor{tabblue}{HTML}{1f77b4}
\definecolor{taborange}{HTML}{ff7f0e}
\definecolor{tabgreen}{HTML}{2ca02c}
\definecolor{tabred}{HTML}{d62728}
\definecolor{tabpurple}{HTML}{9467bd}

\definecolor{cblue}{RGB}{173, 201, 233}
\definecolor{clblue}{RGB}{222, 234, 246}
\definecolor{corange}{RGB}{255, 152, 67}
\definecolor{lorgange}{RGB}{255, 221, 149}
\definecolor{myred}{RGB}{174,66,38}

\newcolumntype{C}{>{\centering\arraybackslash}X}
\newcommand*{\myparagraphnospace}[1]{\noindent\textbf{#1}\hspace{2mm}}

\newcommand{\cc}[1]{\cellcolor{clblue!50}{#1}}

\definecolor{cvprblue}{rgb}{0.21,0.49,0.74}
\usepackage[pagebackref,breaklinks,colorlinks,allcolors=cvprblue]{hyperref}

\def\paperID{****} 
\def\confName{CVPR}
\def\confYear{2025}

\title{Complexity Experts are Task-Discriminative Learners for Any Image Restoration}

\author{Eduard Zamfir$^{1}$ \quad Zongwei Wu$^{1}$\thanks{Corresponding author} \quad Nancy Mehta$^{1}$ \quad Yuedong Tan$^{1}$ \\
        Danda Pani Paudel$^{2}$ \quad Yulun Zhang$^{3}$ \quad Radu Timofte$^{1}$ \\
        \small{$^{1}$University of Würzburg} \quad 
        \small{$^{2}$INSAIT, Sofia University ``St. Kliment Ohridski''} \quad
        \small{$^{3}$Shanghai Jiao Tong University} \\
}

\begin{document}

\maketitle 

\begin{abstract}
Recent advancements in all-in-one image restoration models have revolutionized the ability to address diverse degradations through a unified framework. However, parameters tied to specific tasks often remain inactive for other tasks, making mixture-of-experts (MoE) architectures a natural extension. Despite this, MoEs often show inconsistent behavior, with some experts unexpectedly generalizing across tasks while others struggle within their intended scope. This hinders leveraging MoEs' computational benefits by bypassing irrelevant experts during inference.
We attribute this undesired behavior to the uniform and rigid architecture of traditional MoEs. To address this, we introduce ``complexity experts" -- flexible expert blocks with varying computational complexity and receptive fields. A key challenge is assigning tasks to each expert, as degradation complexity is unknown in advance. Thus, we execute tasks with a simple bias toward lower complexity.
To our surprise, this preference effectively drives task-specific allocation, assigning tasks to experts with the appropriate complexity. 
Extensive experiments validate our approach, demonstrating the ability to bypass irrelevant experts during inference while maintaining superior performance. The proposed MoCE-IR model outperforms state-of-the-art methods, affirming its efficiency and practical applicability.
The source code and models are publicly available at \href{https://eduardzamfir.github.io/moceir/}{\texttt{eduardzamfir.github.io/MoCE-IR/}}
\end{abstract}

\section{Introduction}
\label{sec:intro}

 \begin{figure}[t]
    \centering
    \includegraphics[width=\columnwidth]{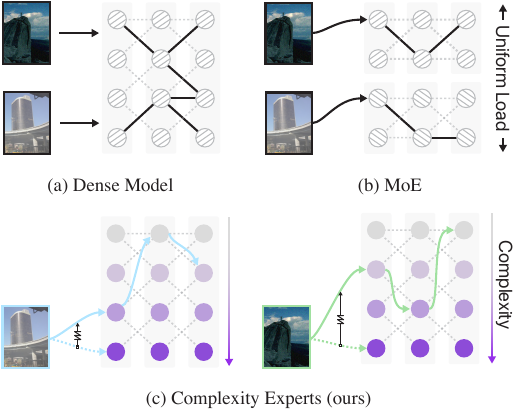}
    \vspace{-6mm}
    \caption{\textit{Motivation.} (a) Dense all-in-one restoration methods~\cite{li2022airnet,yang2024ldr} often inefficiently allocate parameters when handling multiple degradation types.
    (b) While recent Mixture-of-Experts (MoE) approaches~\cite{zamfir2024efficient,yu2024multi} address this through sparse computation, their rigid routing mechanisms uniformly distribute inputs across experts without considering the natural relationships between degradations. 
    (c) To overcome these limitations, we introduce Complexity Experts - adaptive processing blocks with size-varying computational units. Our framework dynamically allocates model capacity using a spring-inspired force mechanism that continuously guides routing decisions toward simpler experts when possible, with the force proportional to the complexity of the input degradation. While initially designed for computational efficiency, this approach naturally emerges as a task-discriminative learning framework, assigning degradations to the most suitable experts. This makes it particularly effective for all-in-one restoration methods, where both task-specific processing and cross-degradation knowledge sharing are crucial.}
    \label{fig:enter-label}
    \vspace{-6mm}
\end{figure}

Image restoration~\cite{zhang2017learning,zhang2019residual,tai2017memnet} is a fundamental problem in computer vision, dealing with reconstructing high-quality images from deteriorated observations. Adverse conditions such as noise~\cite{lehtinen2018noise2noise,zhang2017learning}, haze~\cite{cai2016dehazenet,qu2019enhanced}, or rain~\cite{yasarla2019uncertainty,chen2023alwayscleardays} significantly impact the practical utility of images in downstream applications across various domains, including autonomous navigation~\cite{valanarasu2022transweather,chen2023alwayscleardays} or augmented reality~\cite{girbacia2013virtual,saggio2011augmented,dang2020application}.
Deep learning-based approaches have remarkably advanced this field, particularly for task-specific image restoration problems~\cite{liu2021swin,chen2022simple,wang2022uformer,Zamir2021Restormer,chen2022cross,guo2024mambair}.
Recent all-in-onerestoration models~\cite{li2022airnet,valanarasu2022transweather,liu2022tape,fan2019dl,potlapalli2023promptir,zhang2023ingredient,conde2024high,duan2024uniprocessor}, however, demonstrate the possibility of handling multiple degradation types within a single model, offering more practical solutions compared to traditional task-specific approaches without the extensive re-training need for each new degradation type.
Noteable works employ visual~\cite{potlapalli2023promptir,li2023pip,wang2023promptrestorer} or language-based prompting~\cite{conde2024high,luo2023daclip,ai2024multimodal}, contrastive learning~\cite{li2022airnet,zhang2023ingredient}, and diffusion-based models~\cite{luo2023daclip,ai2024multimodal}.

Despite their success, we observe that the aforementioned models often suffer from inefficiencies, as parameters tied to specific degradations often remain inactive or underutilized when addressing unrelated tasks~\cite{yang2024ldr}.
This naturally suggests the use of mixture-of-experts (MoE) architectures for task-specific processing. Yet, current MoE-based approaches~\cite{yang2024ldr, luo2023wm, zamfir2024efficient} typically incorporate routing mechanisms based on language~\cite{yang2024ldr} or degradation priors~\cite{luo2023wm, zamfir2024efficient}, leading to imbalanced optimization, where some experts generalize well, while others struggle with their intended tasks. This inconsistency limits the potential computational benefits of bypassing task-irrelevant experts during inference.

We attribute this undesired behavior to two limitations in current approaches. First, the uniform and rigid architecture of existing MoE models fails to account for the varying complexity requirements across different restoration tasks. For instance, motion blur demands localized processing with strong spatial awareness, while haze removal requires broader contextual understanding~\cite{pan2016blind,berman2016non}, necessitating adaptive processing aligned to the task requirements. Second, the challenge of appropriately routing tasks to experts is complicated by the unknown complexity of each degradation type a priori. Typically, MoE models~\cite{riquelme2021scaling,puigcerver2024softmoe,raposo2024mod,zhou2022mixture} aim to balance expert utilization, preventing the model from collapsing into a single-expert dependency.

In this work, we introduce a novel mixture-of-complexity-experts (MoCE) framework for all-in-one image restoration that directly addresses these limitations. Our key innovation lies in designing expert blocks with increasing computational complexity and receptive fields, allowing the model to adaptively match processing capacity with task requirements. 

To address the routing challenge, we introduce a complexity-aware allocation mechanism that preferentially directs tasks to lower-complexity experts. This approach implements a complexity-proportional bias term, analogous to a mechanical spring force, to guide expert selection.
Surprisingly, this straightforward approach results in effective task-specific allocation, naturally directing inputs to experts with appropriate complexity levels.
Our MoCE framework offers two significant advantages: it ensures efficient inference by selectively bypassing irrelevant experts, thereby reducing computational overhead, while simultaneously maintaining high-quality restoration performance across diverse tasks. Through extensive experiments across multiple image restoration tasks, we demonstrate that our approach not only preserves restoration quality but surpasses recent state-of-the-art methods. These results validate both the effectiveness of our complexity-based expert design and the practicality of our routing strategy for real-world applications.
Our approach establishes a new benchmark for all-in-one image restoration, enhancing both efficiency and fidelity. 

The principal contributions of this work are threefold:
\begin{itemize}
    \item We propose MoCE-IR, achieving SoTA all-in-one image restoration performance with improved efficiency compared to prior works.
    \item Our novel MoE layer selectively activates complexity experts based on input requirements, unifying task-specific and holistic learning in a single architecture.
    \item We develop a complexity-aware routing mechanism that balances restoration quality with computational efficiency by adaptive expert allocation.
\end{itemize}
\section{Related Works}
\label{sec:related_works}

\begin{figure*}[t]
    \begin{subfigure}{0.33\textwidth}
        \includegraphics[width=\textwidth]{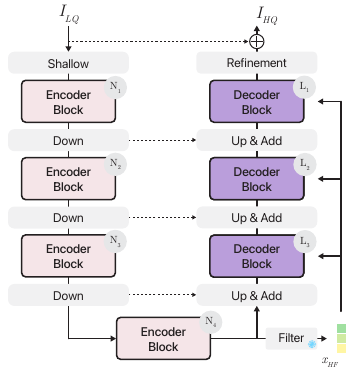}
        \subcaption{\textit{Architecture overview}}
        \label{fig:method:overview}
    \end{subfigure}
    \hfill
    \begin{subfigure}{0.33\textwidth}
        \includegraphics[width=\textwidth]{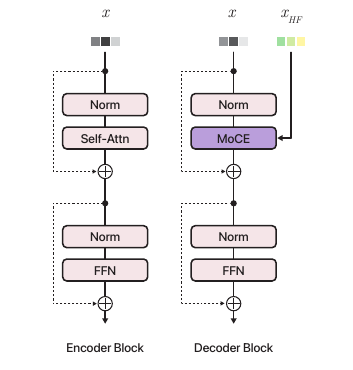}
        \subcaption{\textit{Transformer blocks.}}
        \label{fig:method:blocks}
    \end{subfigure}
    \hfill
    \begin{subfigure}{0.33\textwidth}
        \includegraphics[width=\textwidth]{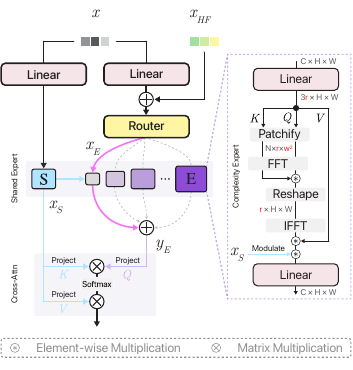}
        \subcaption{\textit{Mixture-of-complexity-experts}}
        \label{fig:method:moce}
    \end{subfigure}
    \vspace{-5mm}
    \caption{\textit{Proposed MoCE-IR framework.} Despite recent advances in MoE-based image restoration~\cite{yang2024ldr,zamfir2024details,zamfir2024efficient}, inconsistent expert behavior—where some experts over-generalize while others underperform—limits their computational efficiency. We address this through \textit{complexity experts}: flexible blocks with varying computational capacity and receptive fields. Our MoCE-IR employs an asymmetric encoder-decoder architecture where each decoder block contains a mixture-of-complexity-experts layer for adaptive capacity routing.}
    \label{fig:method:main}
    \vspace{-5mm}
\end{figure*}

\paragraph{Task Specific Image Restoration.}
Reconstructing the clean image from its degraded counterpart is a highly ill-posed problem, however, a great body of work have addressed image restoration from a data-driven learning perspective, achieving tremendous results compared to prior hand-crafted methods~\cite{zhang2017learning,tai2017memnet,lehtinen2018noise2noise,zhang2019residual,liang2021swinir,Zamir2021Restormer,wang2022uformer}. Most proposed solutions build on convolutional~\cite{zhang2017learning,tai2017memnet,zhang2019residual,chen2022simple} or Transformer-based architectures~\cite{liu2021swin,wang2022uformer,Zamir2021Restormer,chen2022cross} addressing single degradation tasks, such as denoising~\cite{zhang2017learning,zhang2019residual,chen2022cross}, dehazing~\cite{ren2020singledehazing,ren2018gated,wu2021contrastivedehazing} or deraining~\cite{jiang2020multi,ren2019progressive}. Contrary to CNN-based networks, Transformer offer strong modeling capabilities for capturing global dependencies, which makes them outstanding image restorers~\cite{liu2021swin,Zamir2021Restormer,zhao2023comprehensive}. Self-attention's quadratic complexity w.r.t the image size poses a challenge for resource-constrained applications. 
Conversely, convolutions offer fast and efficient processing with limited global context, but they scale more effectively with increasing input size. Recently, FFTformer~\cite{kong2023efficient} introduced an efficient approximation of query and key interactions in the Fourier domain, significantly reducing computational demands compared to window-based attention methods for image deblurring. 
Building on this, we incorporate specialized experts into a Transformer-based architecture~\cite{Zamir2021Restormer} to efficiently capture both shared and distinct contextual information, addressing the diverse demands of image restoration tasks.

\vspace{1mm}
\myparagraphnospace{All-in-One Image Restoration.}
Restoring degraded images often requires multiple models, complicating practical implementation, especially when images exhibit a combination of various degradations.
An emerging field known as all-in-one image restoration is advancing in low-level computer vision, utilizing a single deep blind restoration model to tackle multiple degradation types simultaneously~\cite{zamir2021pmrnet,valanarasu2022transweather,chen2023alwayscleardays,jiang2023autodir,potlapalli2023promptir,zhang2023ingredient}. 
The seminal work, AirNet~\cite{li2022airnet} achieves blind all-in-one image restoration using contrastive learning to extract degradation representations, which guide the restoration process. Similarly, IDR~\cite{zhang2023ingredient} employs a meta-learning-based two-stage approach to model degradation through underlying physical principles. Prompt-based learning~\cite{potlapalli2023promptir, wang2023promptrestorer, li2023pip} has also gained traction, with ~\cite{potlapalli2023promptir} introducing tunable prompts to encode degradation-specific information, albeit with a high parameter count. With recent advancements in multimodal integration, such as combining images and language, there has been growing interest in language-guided image restoration. Notable approaches include DA-CLIP~\cite{luo2023daclip}, InstructIR~\cite{conde2024high}, and UniProcessor~\cite{duan2024uniprocessor}. However, these methods often depend on high-quality text prompts and typically require LLMs, limiting their applicability on resource-constrained devices. In contrast, our contribution lies in a novel, parameter-efficient approach that focuses on selective activation, effectively addressing diverse degradations while remaining lightweight and adaptable. 

\vspace{1mm}
\myparagraphnospace{Dynamic Networks.}
Dynamic networks have evolved from basic conditional computation~\cite{bengio2013estimating} to sophisticated Mixture-of-Experts (MoE) architectures that expand model capacity while maintaining efficient inference costs~\cite{shazeer2017outrageously, riquelme2021scaling, puigcerver2024softmoe}. This approach has proven particularly effective in transformer-based NLP and high-level vision tasks. In image restoration, Path-Restore~\cite{yu2022:pathrestore} introduced content-aware patch routing with difficulty-regulated rewards, while recent all-in-one models~\cite{luo2023wm,yang2024ldr,zamfir2024efficient} leverage various priors for expert routing. Our work addresses the challenge of adaptively routing corrupted images to appropriate computational units within a generalist restoration framework. By dynamically aligning image characteristics with computational requirements, we achieve a unified model in both parameter and architectural space, advancing beyond previous limitations.

\section{Method}
\label{sec:method}

In this section, we outline the core principles of our All-in-One image restoration method. As shown in \cref{fig:method:main}a, our pipeline features a U-shaped architecture~\cite{ronneberger2015unet,Zamir2021Restormer} with an asymmetric encoder-decoder design. 
A $3\times3$ convolution first extracts shallow features from the degraded input, which then pass through $4$ levels of encoding and decoding stages. While both stages utilize Transformer blocks~\cite{Zamir2021Restormer,potlapalli2023promptir}, our novel MoCE layers are exclusively integrated into the decoder blocks (\cref{fig:method:blocks}). To further enhance the decoder's feature enrichment capabilities, we introduce high-frequency guidance via Sobel-filtered global feature vector to promote the frequency awareness of the gating functions. 
Finally, a global residual connection links the shallow features to the output of the refinement stage, refining the crucial high-frequency details before producing the restored image.

\subsection{Mixture-of-Complexity-Experts}
\label{sec:method:moce}
To overcome the limitations of uniform MoE architectures in image restoration, we propose an efficient input-adaptive model for all-in-one image restoration that dynamically adjusts processing capacity to match task requirements.

Technically, at the core of our framework is a specialized MoCE layer, consisting of $n$ complexity experts $\mathbf{E}$ and a single shared expert $\mathbf{S}$. As shown in \cref{fig:method:blocks}, our experts are designed with increasing computational complexity and progressively larger receptive fields, in contrast to the uniform architecture of traditional MoE designs~\cite{shazeer2017outrageously,riquelme2021scaling,puigcerver2024softmoe}. 
The interaction between these components is orchestrated through a two-level gating mechanism: first projecting routed tokens into expert-specific embeddings, then combining specialized and agnostic features through modulation, and finally merging the outputs via cross-attention. This enables our model to capture both degradation-specific features and inter-degradation relationships, while maintaining a bias toward computationally efficient processing paths.

\vspace{1mm}
\myparagraphnospace{Expert Design.}
Each complexity expert block as shown in \cref{fig:method:moce} is designed based on two fundamental principles: ensuring computational efficiency and capturing hierarchical spatial features, which are critical for restoring fine details and addressing diverse degradation patterns. Given the growing computational demands associated with increasing expert numbers and capacities, it is crucial to prioritize an overall efficient expert design. 
To this end, we implement a nested expert structure, progressively reducing channel dimensionality \textcolor{myred}{$r$} within each subsequent expert to control the computational overhead.
Simultaneously, we increase the receptive field, \textit{i.e.} window partition size \textcolor{myred}{$w$}, within each expert to adaptively balance the localized and global processing, tailored to the specific requirements of the input degradation. Thus, our complexity experts are constructed by scaling the design along two key axes: window partition size and channel dimension.

More specifically, each expert $\mathbf{E}$ projects its input tokens to the embedding \(\mathbf{x}_{E} \in \mathbb{R}^{H \times W \times \textcolor{myred}{r}}\), where \( \textcolor{myred}{r} = \nicefrac{C}{2^{i}}~\text{for}~i \in \{i,...,n\} \), to highlight the most relevant information along the channel dimension. 
Following this, a window-based self-attention (WSA) mechanism captures spatial information effectively. 
To optimize performance, we employ an FFT-based approximation~\cite{kong2023efficient} for efficient matrix multiplication between 
queries \(Q\) and keys \(K\) in the Fourier domain. 
The refined features are then transformed back into the spatial domain and re-projected to their original dimensionality. 
Lastly, the output of the expert block $\mathbf{\hat{x}}_{E}^{i}$ is modulated through element-wise multiplication with the holistic features from the shared expert, $\mathbf{S}$. As shared expert, we employ the transposed self-attention (T-SA) module~\cite{Zamir2021Restormer} in the channel dimension, while the complexity experts operate spatially.
More concretely, the modulated expert features $\mathbf{y_{E}^{i}}$ are obtained as following:
\begin{align}
    \mathbf{y_{E}^{i}} &=\mathbf{\hat{x}}_{E}^{i} \odot \mathbf{S}(\mathbf{x})\\
    \text{with}~ \mathbf{\hat{x}}_{E}^{i} &= \mathbf{E}^{i}(\mathbf{x}_{E}^{i}) = \text{FFT-WSA}_{\textcolor{myred}{\text{w}_{i}}}(\mathbf{W}_{C \rightarrow r_{i}}^{i} \mathbf{x}_{E}^{i}) \\
    \text{and}~ \mathbf{S}(\mathbf{x})&= \text{T-SA}(\mathbf{x})
\end{align}
where linear layers for projection are denoted as $\mathbf{W}$, $\odot$ denotes element-wise multiplication and $i$ is the index of the current complexity expert $\mathbf{E}^{i}$ with window size \textcolor{myred}{$w_{i}$} and embedding dimensionality \textcolor{myred}{$r_{i}$}.
For example, the most lightweight expert is characterized by the smallest embedding dimension \textcolor{myred}{\(r_{1}\)} and window size \textcolor{myred}{\(w_{1}\)}, while the dimensions and window partitions for the other experts can increase linearly. 
Throughout our network, the capacity of the shared expert remains constant, with \(\textcolor{myred}{r}=C\).

\subsection{Complexity-aware Routing}
\label{sec:method:complexity}
Drawing inspiration from sparse MoE~\cite{shazeer2017outrageously,riquelme2021scaling}, we integrate linear layers within each decoder block to enable a routing mechanism, associating input features $\mathbf{x} \in \mathbb{R}^{H \times W \times C}$ with their corresponding  specialized complexity experts $\mathbf{E}$. 
However, in the context of image restoration, a key challenge arises in achieving a scale-invariant tokenization of input images, ensuring consistency across varying resolutions and scales. While previous MoE approaches~\cite{shazeer2017outrageously,riquelme2021scaling,puigcerver2024softmoe} employ a token-based routing, we opt for an image-level routing strategy instead, where we select experts for  entire input image.
The routing function $g(\mathbf{x})$ orchestrates the allocation of input samples based on the required compute to corresponding complexity experts $\ \mathbf{E}^{i}$, indexed by $i\in 1,...,n$ with $n$ representing the total number of experts. Within $g(\mathbf{x})$, we select the $\text{top-}{k}$ elements of the softmax distribution associating input tokens to experts, setting all other elements to zero. In practice, we use $k=1$ while $\epsilon$ is sampled independently $\epsilon \sim \mathcal{N}(0, \nicefrac{1}{n^2})$ entry-wise enabling noisy $\text{top-}{1}$ routing:
\begin{align}
    g(\mathbf{x}) = \text{top}_{k}(\text{Softmax}(\mathbf{Wx}+\epsilon))
\end{align}
During training, we incorporate an auxiliary loss \(\mathcal{L}_{\text{aux}}\) inspired by the load balancing loss proposed by Riquelme \textit{et al.}~\cite{riquelme2021scaling}. We complement the importance term to favor experts based on their computational complexity, allowing for more effective balancing aligned with expert capacity.
As in \cite{riquelme2021scaling}, the balanced expert utilization is encouraged through an importance loss, where the importance of each expert \(\mathbf{E}^{i}\) for a batch of images is defined as the sum of the routing weights associated with the \(i\)-th expert across the entire batch. Additionally, we introduce a complexity bias \(\mathbf{b}\) by calculating the number of learnable parameters \(p_{i}\) in each expert \(\mathbf{E}^{i}\) and normalizing these values with respect to the largest parameter count of the experts, \(p_{\text{max}}\):
\begin{align}   
    \text{Imp}_{i}(\mathbf{x}) &= \left( \sum_{x \in \mathbf{x}} \text{Softmax}(Wx)_{i} \right) \ast \mathbf{b}, \\
    \text{with}~\mathbf{b} &= \left[ \nicefrac{p_1}{p_{\text{max}}}, \nicefrac{p_2}{p_{\text{max}}}, \dots, \nicefrac{p_n}{p_{\text{max}}} \right],
\end{align}
where $W$ is the layer-specific weight for the router $g(\mathbf{x})$. This approach assigns progressively lower weights to experts with lower complexity. For intuitive understanding, this mechanism parallels a mechanical spring system, where the restoring force is proportional to both displacement and the spring constant. In our context, the expert's parameter count corresponds to displacement, while the normalization factor serves as the material-specific constant, collectively determining the magnitude of complexity bias. Without this, the routers lack a basis for selecting specific experts, leading to randomized routing and thus suboptimal model capacity utilization. Lastly, the complexity-aware importance \(\text{Imp}\) is used to compute the total auxiliary loss \(\mathcal{L}_{\text{aux}}\) defined as:
\begin{align}
\mathcal{L}_{\text{aux}}(\mathbf{x}) = \nicefrac{1}{2}~\text{CV}(\text{Imp}(\mathbf{x}))^{2} + \nicefrac{1}{2}~\text{CV}(\text{Load}(\mathbf{x}))^{2},
\end{align}
where CV denotes the coefficient of variation~\cite{riquelme2021scaling}. The effect of the auxiliary loss is shown in \cref{fig:exp:routing_aio3} and \cref{tab:exp:expert_design}.

\section{Experiments}
\label{sec:experiments}

\begin{table*}[t]
    \centering
    \scriptsize
    \fboxsep0.75pt
    \setlength\tabcolsep{7pt}
    \caption{\textit{Comparison to state-of-the-art on three degradations.} PSNR (dB, $\uparrow$) and \colorbox{clblue!50}{SSIM ($\uparrow$)} are reported on the full RGB images. Our method sets a new state-of-the-art on average across all benchmarks on two model scales, showcasing its scalability while being more efficient than prior work. ‘-’ represents unreported results and ‘$\ast$’ indicates VLM-based approaches. The \textbf{best} performances are highlighted.}
    \vspace{-3mm}
    \label{tab:exp:3deg}
    \begin{tabularx}{\textwidth}{lX*{15}{c}}
    \toprule
     & \multirow{2}{*}{Method} & \multirow{2}{*}{Params.} 
     & \multicolumn{2}{c}{\textit{Dehazing}} & \multicolumn{2}{c}{\textit{Deraining}} & \multicolumn{6}{c}{\textit{Denoising}} 
     & \multicolumn{2}{c}{\multirow{2}{*}{Average}}\\
     \cmidrule(lr){4-5} \cmidrule(lr){6-7} \cmidrule(lr){8-13} 
     &&& \multicolumn{2}{c}{SOTS} & \multicolumn{2}{c}{Rain100L} & \multicolumn{2}{c}{BSD68\textsubscript{$\sigma$=15}} & \multicolumn{2}{c}{BSD68\textsubscript{$\sigma$=25}} & \multicolumn{2}{c}{BSD68\textsubscript{$\sigma$=50}}\\
     \midrule
     \multirow{6}{*}{\rotatebox{90}{Light}} &
         BRDNet~\cite{tian2000brdnet} & - & 23.23 & \cc{.895} & 27.42 & \cc{.895} & 32.26 & \cc{.898} & 29.76 & \cc{.836} & 26.34 & \cc{.693} & 27.80 & \cc{.843} \\
        & LPNet~\cite{gao2019dynamic} & - & 20.84 & \cc{828} & 24.88 & \cc{.784} & 26.47 & \cc{.778} & 24.77 & \cc{.748} & 21.26 & \cc{.552} & 23.64 & \cc{.738} \\
        & FDGAN~\cite{dong2020fdgan}  & - & 24.71 & \cc{.929} & 29.89 & \cc{.933} & 30.25 & \cc{.910} & 28.81 & \cc{.868} & 26.43 & \cc{.776} & 28.02 & \cc{.883} \\
         & DL~\cite{fan2019dl} & 2M & 26.92 &\cc{.931} & 32.62 &\cc{.931} & 33.05 &\cc{.914} & 30.41 &\cc{.861} & 26.90 &\cc{.740}  & 29.98 &\cc{.876}\\
        & AirNet~\cite{li2022airnet} & 9M & 27.94 & \cc{.962} & 34.90 & \cc{.967} & 33.92 &\cc{.933} & 31.26 & \cc{.888} & 28.00 & \cc{.797} & 31.20 & \cc{.910} \\
        & MoCE-IR-S (\textit{ours}) & 11M &\textbf{30.94}&\cc{\textbf{.979}} & \textbf{38.22} &\cc{\textbf{.983}} & \textbf{34.08} &\cc{\textbf{.933}} & \textbf{31.42} &\cc{\textbf{.888}} & \textbf{28.16} &\cc{\textbf{.798}} & \textbf{32.57} &\cc{\textbf{.916}} \\
        \midrule
    \multirow{7}{*}{\rotatebox{90}{Heavy}} &
         MPRNet~\cite{zamir2021pmrnet}  & 16M & 25.28 & \cc{.955} & 33.57 & \cc{.954} & 33.54 & \cc{.927} & 30.89 & \cc{.880} & 27.56 & \cc{.779} & 30.17 & \cc{.899} \\
        & PromptIR~\cite{potlapalli2023promptir} & 36M &30.58 & \cc{.974} & 36.37 &\cc{.972} & 33.98 & \cc{.933} & 31.31 &\cc{.888} & 28.06 & \cc{.799} & 32.06 & \cc{.913} \\
        & Gridformer~\cite{wang2024gridformer} & 34M & 30.37 & \cc{.970} & 37.15 & \cc{.972} & 33.93 &\cc{.931} & 31.37 &\cc{.887} & 28.11 &\cc{.801} & 32.19 &\cc{.912}  \\
        & Art-PromptIR~\cite{wu2024art} & 33M & 30.83 &\cc{.979} & 37.94 &\cc{.982} & 34.06 &\cc{.934} & 31.42 &\cc{.891} & 28.14 &\cc{.801} & 32.49 &\cc{.917} \\
        & DA-CLIP$^\ast$~\cite{luo2023daclip} & 125M & 29.46 &\cc{.963} & 36.28 &\cc{.968} & 30.02 &\cc{.821} & 24.86 &\cc{.585} & 22.29 &\cc{.476} & - &\cc{-}\\
        & UniProcessor$^\ast$~\cite{duan2024uniprocessor} & 1002M & \textbf{31.66} & \cc{.979} & 38.17 & \cc{.982} & 34.08 &\cc{\textbf{.935}} & 31.42 & \cc{\textbf{.891}} & 28.17 & \cc{\textbf{.803}} & 32.70 & \cc{\textbf{.918}}\\
        & MoCE-IR~(\textit{ours}) & 25M & 31.34 & \cc{\textbf{.979}} & \textbf{38.57} & \cc{\textbf{.984}} & \textbf{34.11} & \cc{.932} & \textbf{31.45} & \cc{.888} & \textbf{28.18}  & \cc{.800} & \textbf{32.73}& \cc{.917}\\ 

     \bottomrule
    \end{tabularx}
    \vspace{-3mm}
\end{table*}

\begin{table*}[t]
    \centering
    \scriptsize
    \fboxsep0.75pt
    \setlength\tabcolsep{7pt}
    \caption{\textit{Comparison to state-of-the-art on five degradations.} PSNR (dB, $\uparrow$) and \colorbox{clblue!50}{SSIM ($\uparrow$)} are reported on the full RGB images with $(^\ast)$ denoting general image restorers, others are specialized all-in-one approaches. Under more challenging degradations, our MoCE-IR framework outperforms prior work across two model scales. The \textbf{best} performances are highlighted.}
    \vspace{-3mm}
    \label{tab:exp:5deg}
    \begin{tabularx}{\textwidth}{lX*{15}{c}}
    \toprule
     & \multirow{2}{*}{Method} & \multirow{2}{*}{Params.} 
     & \multicolumn{2}{c}{\textit{Dehazing}} & \multicolumn{2}{c}{\textit{Deraining}} & \multicolumn{2}{c}{\textit{Denoising}} 
     & \multicolumn{2}{c}{\textit{Deblurring}} & \multicolumn{2}{c}{\textit{Low-Light}} & \multicolumn{2}{c}{\multirow{2}{*}{Average}}  \\
     \cmidrule(lr){4-5} \cmidrule(lr){6-7} \cmidrule(lr){8-9} \cmidrule(lr){10-11} \cmidrule(lr){12-13}
     &&& \multicolumn{2}{c}{SOTS} & \multicolumn{2}{c}{Rain100L} & \multicolumn{2}{c}{BSD68\textsubscript{$\sigma$=25}} 
     & \multicolumn{2}{c}{GoPro} & \multicolumn{2}{c}{LOLv1} &  \\
     \midrule
    \multirow{5}{*}{\rotatebox{90}{Light}} &
    SwinIR$^\ast$~\cite{liang2021swinir} & 1M & 21.50 &\cc{.891} & 30.78 &\cc{.923} & 30.59 & \cc{.868} & 24.52 & \cc{.773} & 17.81 & \cc{.723} & 25.04 & \cc{.835} \\ 
    & DL~\cite{fan2019dl} & 2M & 20.54 & \cc{.826} & 21.96 & \cc{.762} & 23.09 & \cc{.745} & 19.86 &\cc{.672} & 19.83 & \cc{.712} & 21.05 & \cc{.743} \\
    & TAPE~\cite{liu2022tape} & 1M & 22.16 & \cc{.861} & 29.67 & \cc{.904} & 30.18 & \cc{.855} & 24.47 & \cc{.763} & 18.97 & \cc{.621} & 25.09 & \cc{.801} \\
    & AirNet~\cite{li2022airnet} & 9M & 21.04 & \cc{.884} & 32.98 & \cc{.951} & 30.91 &\cc{.882} & 24.35 & \cc{.781} & 18.18 & \cc{.735} & 25.49 & \cc{.847} \\
     & MoCE-IR-S~(\textit{ours})& 11M & \textbf{31.33} &\cc{\textbf{.978}} &\textbf{37.21} & \cc{\textbf{.978}} & \textbf{31.25} & \cc{\textbf{.884}} & \textbf{28.90} & \cc{\textbf{.877}} & \textbf{21.68} & \cc{\textbf{.851}} & \textbf{30.08} & \cc{\textbf{.913}}\\
    \midrule
    \multirow{9}{*}{\rotatebox{90}{Heavy}} &
    NAFNet$^\ast$~\cite{chen2022simple} & 17M & 25.23 &\cc{.939} & 35.56 &\cc{.967} & 31.02 & \cc{.883} & 26.53 & \cc{.808} & 20.49 & \cc{.809} & 27.76 & \cc{.881} \\
    & DGUNet$^\ast$~\cite{mou2022dgunet} & 17M & 24.78 &\cc{.940} & 36.62 &\cc{.971} & 31.10 & \cc{.883} & 27.25 & \cc{.837} & 21.87 & \cc{.823} & 28.32 & \cc{.891} \\
    & Restormer$^\ast$~\cite{Zamir2021Restormer} & 26M & 24.09 &\cc{.927} & 34.81 &\cc{.962} & 31.49 & \cc{.884} & 27.22 & \cc{.829} & 20.41 & \cc{.806} & 27.60 & \cc{.881} \\
    & MambaIR~\cite{guo2024mambair} & 27M & 25.81 & \cc{.944} & 36.55 & \cc{.971} & 31.41 & \cc{.884} & 28.61 & \cc{.875} & 22.49 & \cc{.832} & 28.97 & \cc{.901} \\    
    & Transweather~\cite{valanarasu2022transweather} & 38M & 21.32 & \cc{.885} & 29.43 & \cc{.905} & 29.00 & \cc{.841} & 25.12 & \cc{.757} & 21.21 & \cc{.792} & 25.22 & \cc{.836} \\
    & IDR~\cite{zhang2023ingredient} & 15M & 25.24 & \cc{.943} & 35.63 & \cc{.965} &\textbf{31.60} &\cc{.887} & 27.87 &\cc{.846} &21.34 & \cc{.826} & 28.34 & \cc{.893} \\
    & Gridformer~\cite{wang2024gridformer} & 34M & 26.79 & \cc{.951} & 36.61 & \cc{.971} & 31.45 & \cc{.885} & 29.22 & \cc{.884} & 22.59 & \cc{.831} & 29.33 & \cc{.904} \\
    & InstructIR-5D~\cite{conde2024high} & 17M & 27.10 &\cc{.956} & 36.84  &\cc{.973} & 31.40 &\cc{.873} & 29.40 &\cc{.886} & 23.00  &\cc{.836} & 29.55 &\cc{.908} \\
    & MoCE-IR~(\textit{ours}) & 25M & \textbf{30.48} & \cc{\textbf{.974}} & \textbf{38.04} & \cc{\textbf{.982}} & 31.34 & \cc{\textbf{.887}} & \textbf{30.05} & \cc{\textbf{.899}} & \textbf{23.00} & \cc{\textbf{.852}} &\textbf{30.58} &\cc{\textbf{.919}}\\
     \bottomrule
    \end{tabularx}
    \vspace{-3mm}
\end{table*}

\begin{table*}[t]
\centering
\scriptsize
\fboxsep0.75pt
\setlength\tabcolsep{1.55pt}
\caption{\textit{Comparison to state-of-the-art on composited degradations.} PSNR (dB, $\uparrow$) and \colorbox{clblue!50}{SSIM ($\uparrow$)} are reported on the full RGB images with $(^\ast)$ denoting general image restorers, others are specialized all-in-one approaches.  Our MoCE-IR method consistently outperforms even larger models, with favorable results in composited degradation scenarios.}
\vspace{-3mm}
\label{tab:exp:cdd11}
    \begin{tabularx}{\textwidth}{Xc*{8}{c}*{10}{c}*{4}{c}cc}
    \toprule
    \multirow{2}{*}{Method} & \multirow{2}{*}{Params.} & \multicolumn{8}{c}{\textit{CDD11-Single}} & \multicolumn{10}{c}{\textit{CDD11-Double}} & \multicolumn{4}{c}{\textit{CDD11-Triple}} & \multicolumn{2}{c}{\multirow{2}{*}{Avg.}}\\
    \cmidrule(lr){3-10} \cmidrule(lr){11-20} \cmidrule(lr){21-24}
    && \multicolumn{2}{c}{Low~(L)} & \multicolumn{2}{c}{Haze~(H)} & \multicolumn{2}{c}{Rain~(R)} & \multicolumn{2}{c}{Snow~(S)}
    & \multicolumn{2}{c}{L+H} & \multicolumn{2}{c}{L+R} & \multicolumn{2}{c}{L+S} & \multicolumn{2}{c}{H+R} & \multicolumn{2}{c}{H+S} 
    &  \multicolumn{2}{c}{L+H+R} &  \multicolumn{2}{c}{L+H+S} \\ 
    \midrule
    AirNet~\cite{li2022airnet} & 9M
    & 24.83&\cc{.778} & 24.21&\cc{.951} & 26.55&\cc{.891} & 26.79&\cc{.919}
    & 23.23&\cc{.779} & 22.82&\cc{.710} & 23.29&\cc{.723} & 22.21&\cc{.868} & 23.29&\cc{.901}
    & 21.80&\cc{.708} & 22.24&\cc{.725} & 23.75&\cc{.814} \\
    PromptIR~\cite{potlapalli2023promptir} & 36M
    & 26.32&\cc{.805} & 26.10&\cc{.969} & 31.56&\cc{.946} & 31.53&\cc{.960} 
    & 24.49&\cc{.789} & 25.05&\cc{.771} & 24.51&\cc{.761} & 24.54&\cc{.924} & 23.70&\cc{.925} 
    & 23.74&\cc{.752} & 23.33&\cc{.747} & 25.90&\cc{.850} \\
    WGWSNet~\cite{zhu2023Weather} & 26M
    & 24.39&\cc{.774} & 27.90&\cc{.982} & 33.15&\cc{.964} & 34.43&\cc{.973} 
    & 24.27&\cc{.800} & 25.06&\cc{.772} & 24.60&\cc{.765} & 27.23&\cc{.955} & 27.65&\cc{.960}  
    & 23.90&\cc{.772} & 23.97&\cc{.771} & 26.96&\cc{.863} \\
    WeatherDiff~\cite{oezdenizci2022WeatherDiff}  & 83M
    & 23.58&\cc{.763} & 21.99&\cc{.904} & 24.85&\cc{.885} & 24.80&\cc{.888} 
    & 21.83&\cc{.756} & 22.69&\cc{.730} & 22.12&\cc{.707} & 21.25&\cc{.868} & 21.99&\cc{.868} 
    & 21.23&\cc{.716} & 21.04&\cc{.698} & 22.49&\cc{.799} \\
    OneRestore~\cite{guo2024onerestore} & 6M
    & 26.48&\cc{\textbf{.826}} & 32.52&\cc{.990} & 33.40&\cc{.964} & 34.31&\cc{.973}  
    & 25.79&\cc{\textbf{.822}} & 25.58&\cc{.799} & 25.19&\cc{.789} & \textbf{29.99}&\cc{.957} & \textbf{30.21}&\cc{.964} 
    & 24.78&\cc{.788} & 24.90&\cc{\textbf{.791}} & 28.47&\cc{.878} \\
    MoCE-IR-S (\textit{ours}) & 11M 
    & \textbf{27.26}&\cc{.824} & \textbf{32.66}&\cc{\textbf{.990}} & \textbf{34.31}&\cc{\textbf{.970}} & \textbf{35.91}&\cc{\textbf{.980}}
    & \textbf{26.24}&\cc{.817} & \textbf{26.25}&\cc{\textbf{.800}} & \textbf{26.04}&\cc{\textbf{.793}} & 29.93&\cc{\textbf{.964}} & 30.19&\cc{\textbf{.970}}
    & \textbf{25.41}&\cc{\textbf{.789}} &\textbf{25.39}&\cc{.790} & \textbf{29.05}&\cc{\textbf{.881}}
    \\
    \bottomrule
\end{tabularx}
\vspace{-3mm}
\end{table*}

\begin{figure*}[t]
    \centering
    \scriptsize
    \begin{subfigure}{\textwidth}
    \begin{tblr}{
      colspec = {@{}lX[c]X[c]X[c]X[c]X[c]@{}}, colsep=0.01pt, rows={rowsep=0.5pt}, stretch = 0,
    }
     &  Input & AirNet & PromptIR & Ours & GT \\
        \SetCell[r=1]{l}{\rotatebox{90}{\hspace{10mm}Dehazing}} &       
        \includegraphics[width=0.19\textwidth, ]{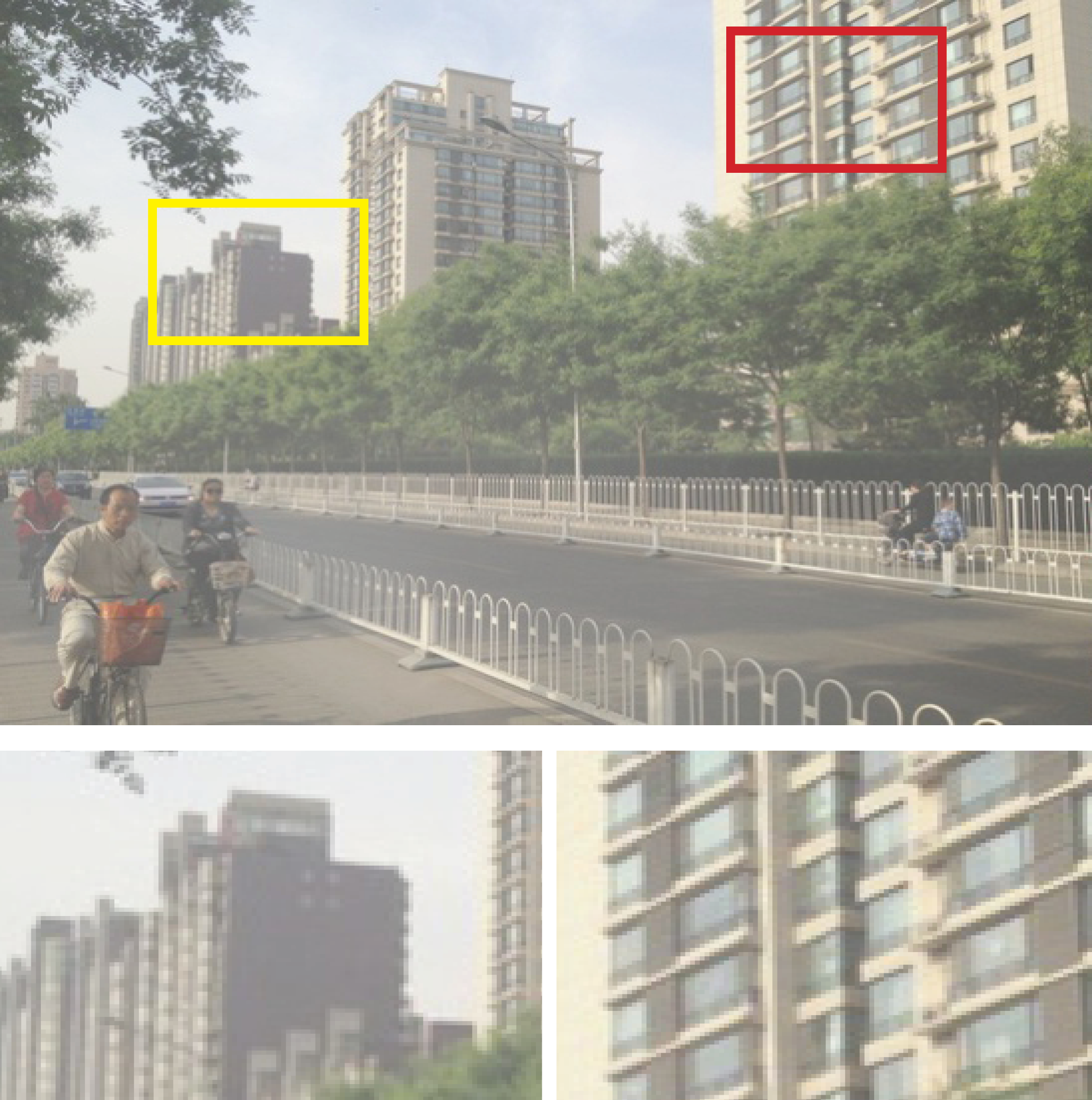} & 
        \includegraphics[width=0.19\textwidth, ]{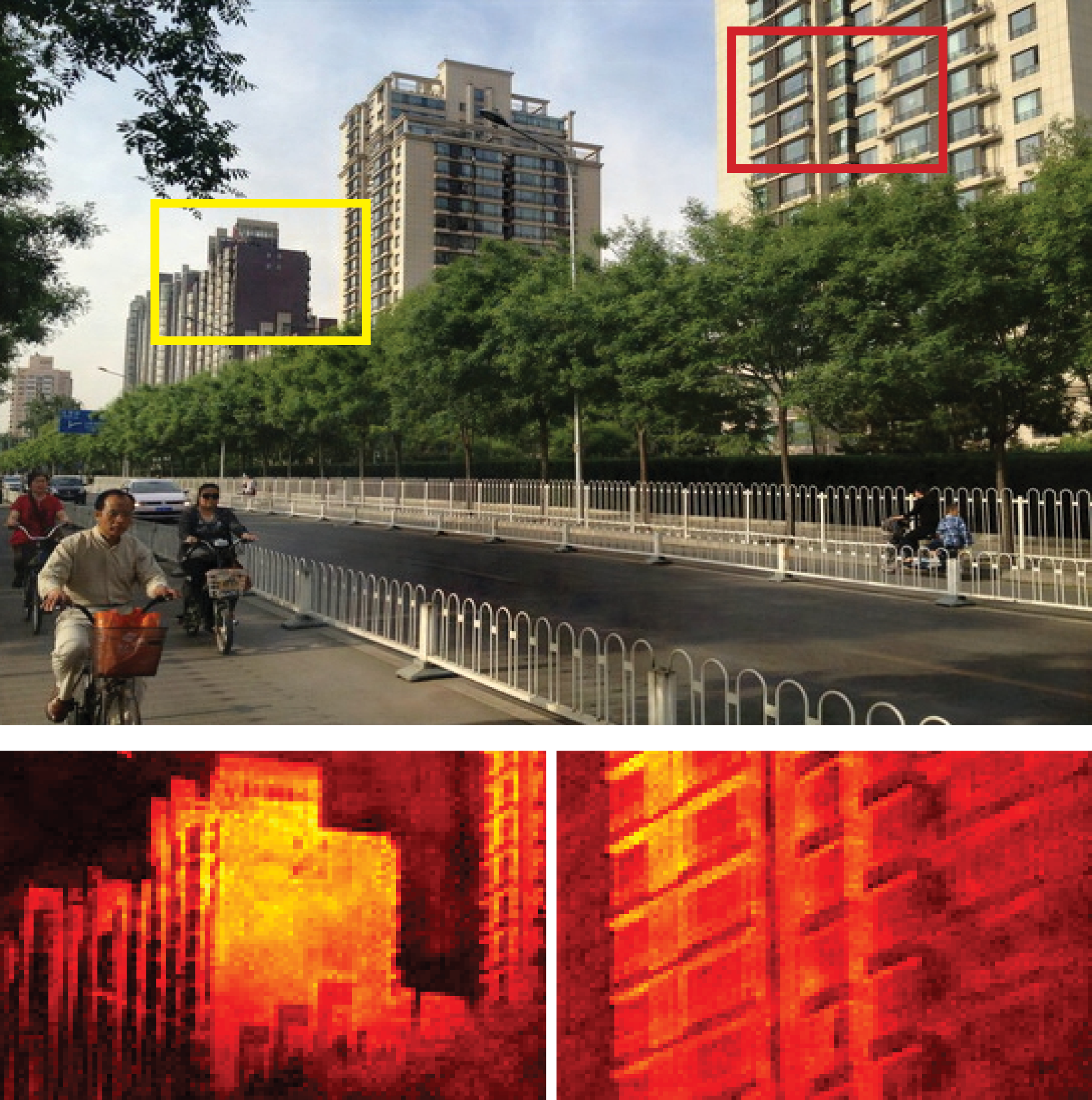} & 
        \includegraphics[width=0.19\textwidth, ]{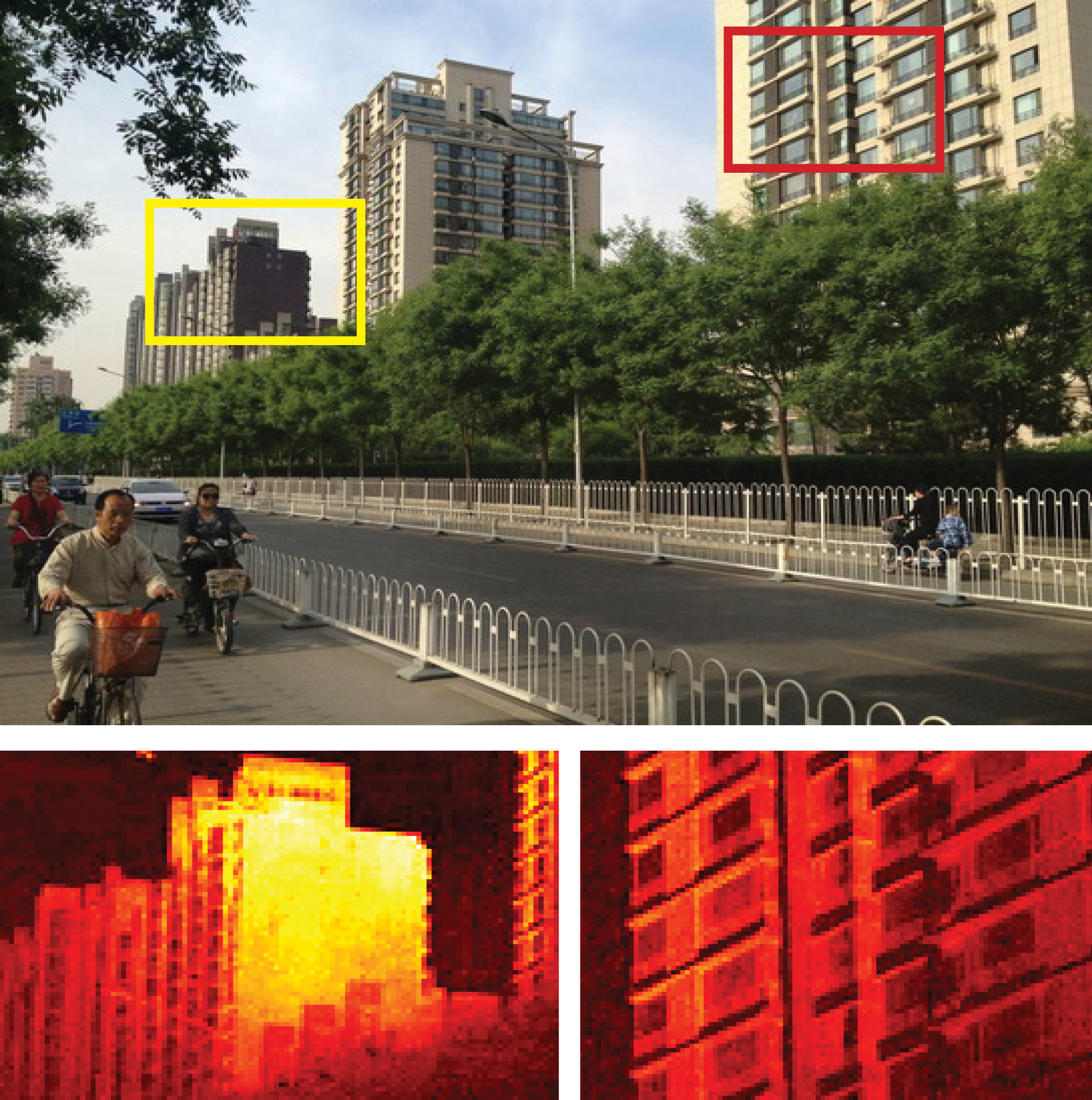} & 
        \includegraphics[width=0.19\textwidth, ]{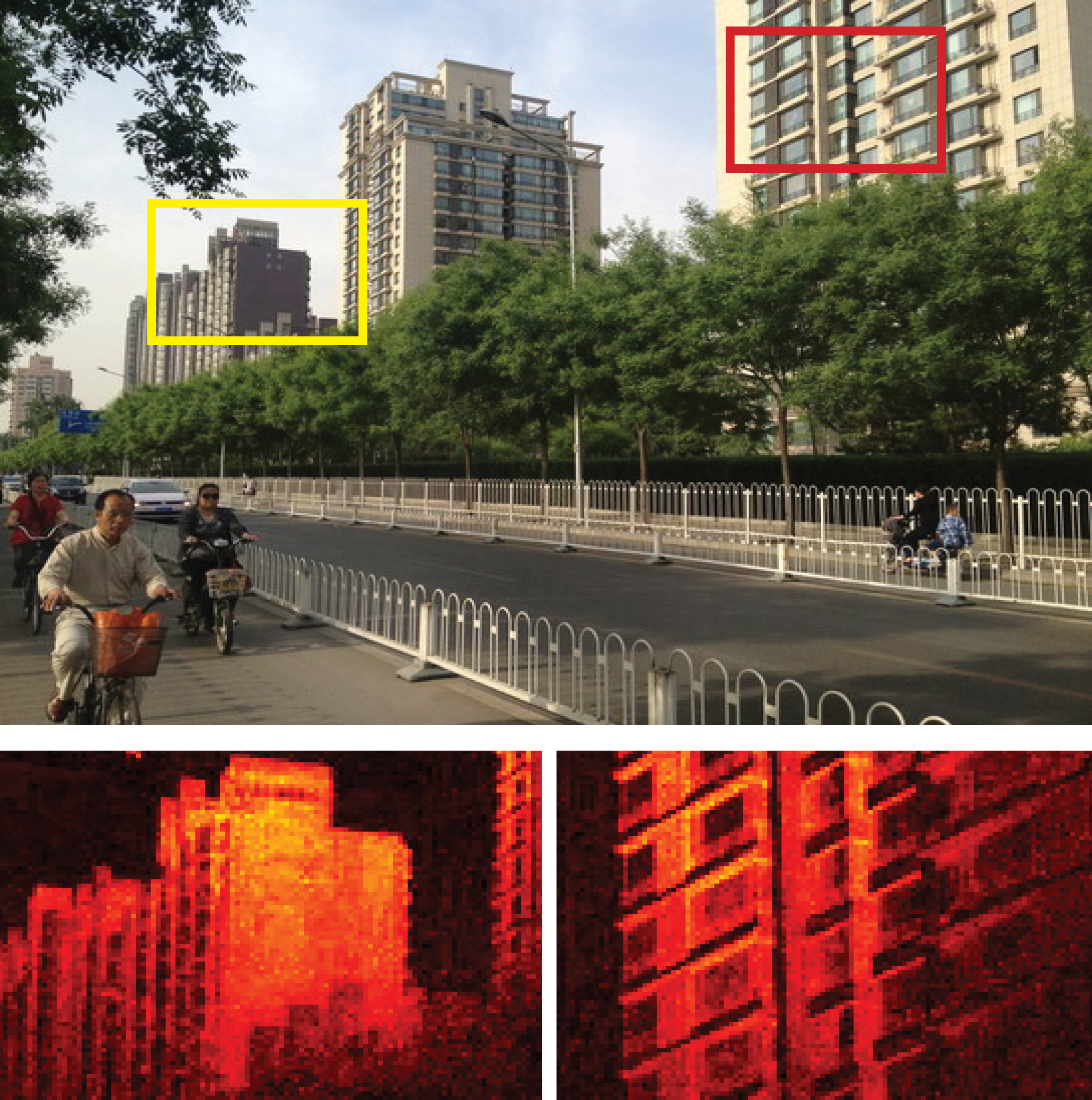} & 
        \includegraphics[width=0.19\textwidth, ]{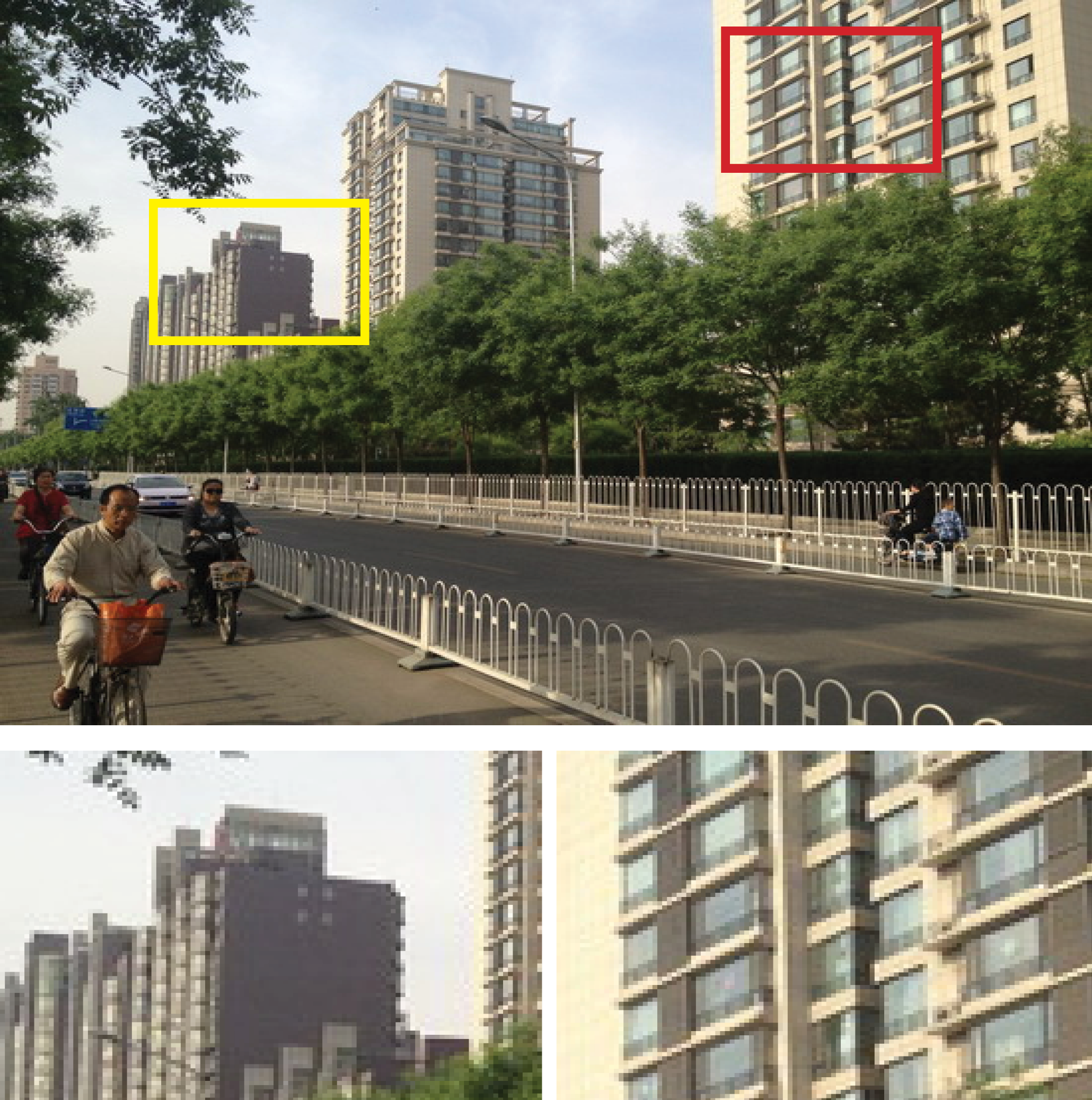}\\
    \end{tblr}
    \end{subfigure}
\hfill
    \begin{subfigure}{\textwidth}
        \begin{tblr}{
      colspec = {@{}lX[c]X[c]X[c]X[c]X[c]@{}}, colsep=0.01pt, rows={rowsep=0.5pt}, stretch = 0,
    }
        \SetCell[r=1]{l}{\rotatebox{90}{\hspace{10mm}Deraining}} & 
        \includegraphics[width=0.19\textwidth, ]{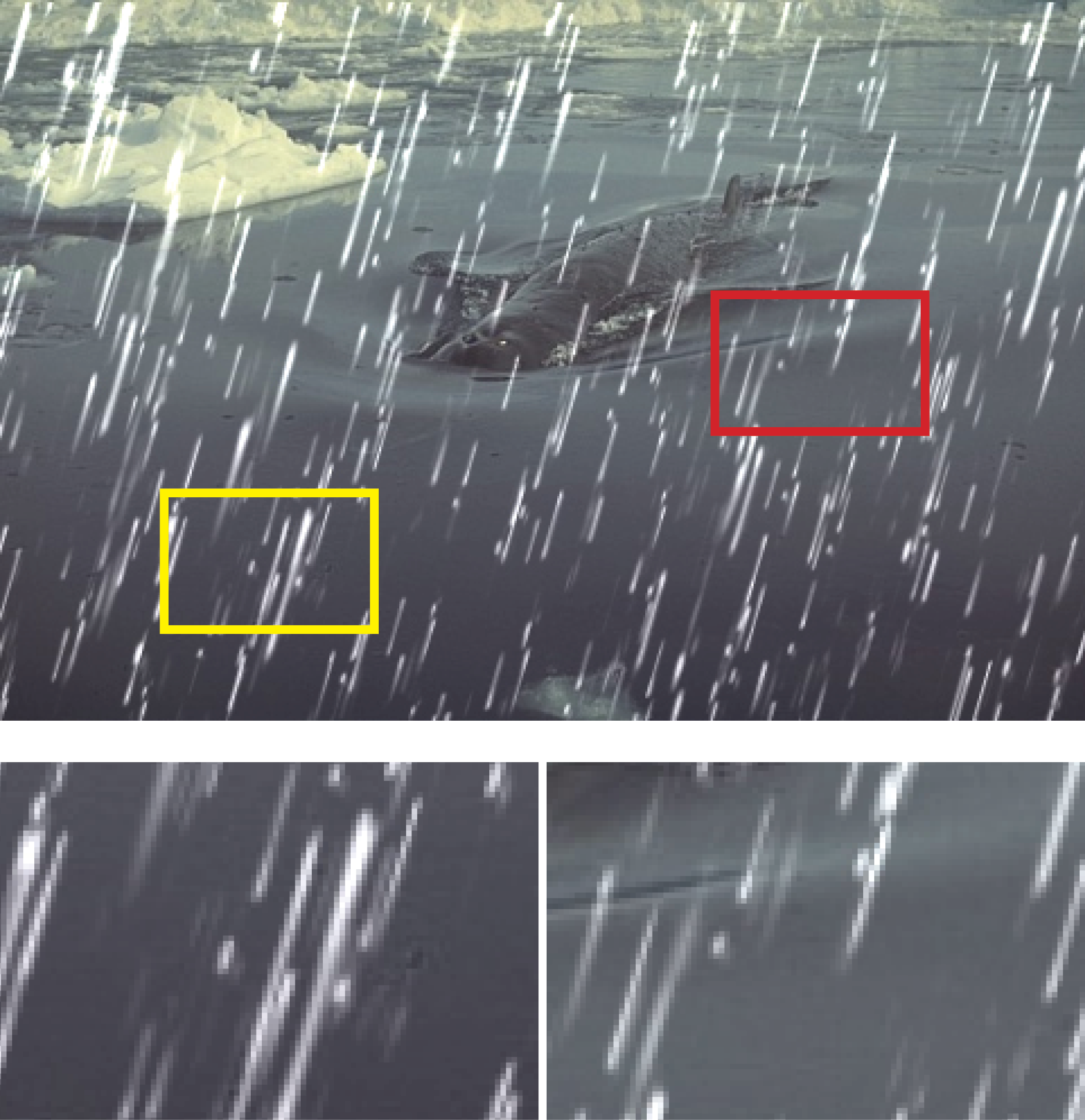} & 
        \includegraphics[width=0.19\textwidth, ]{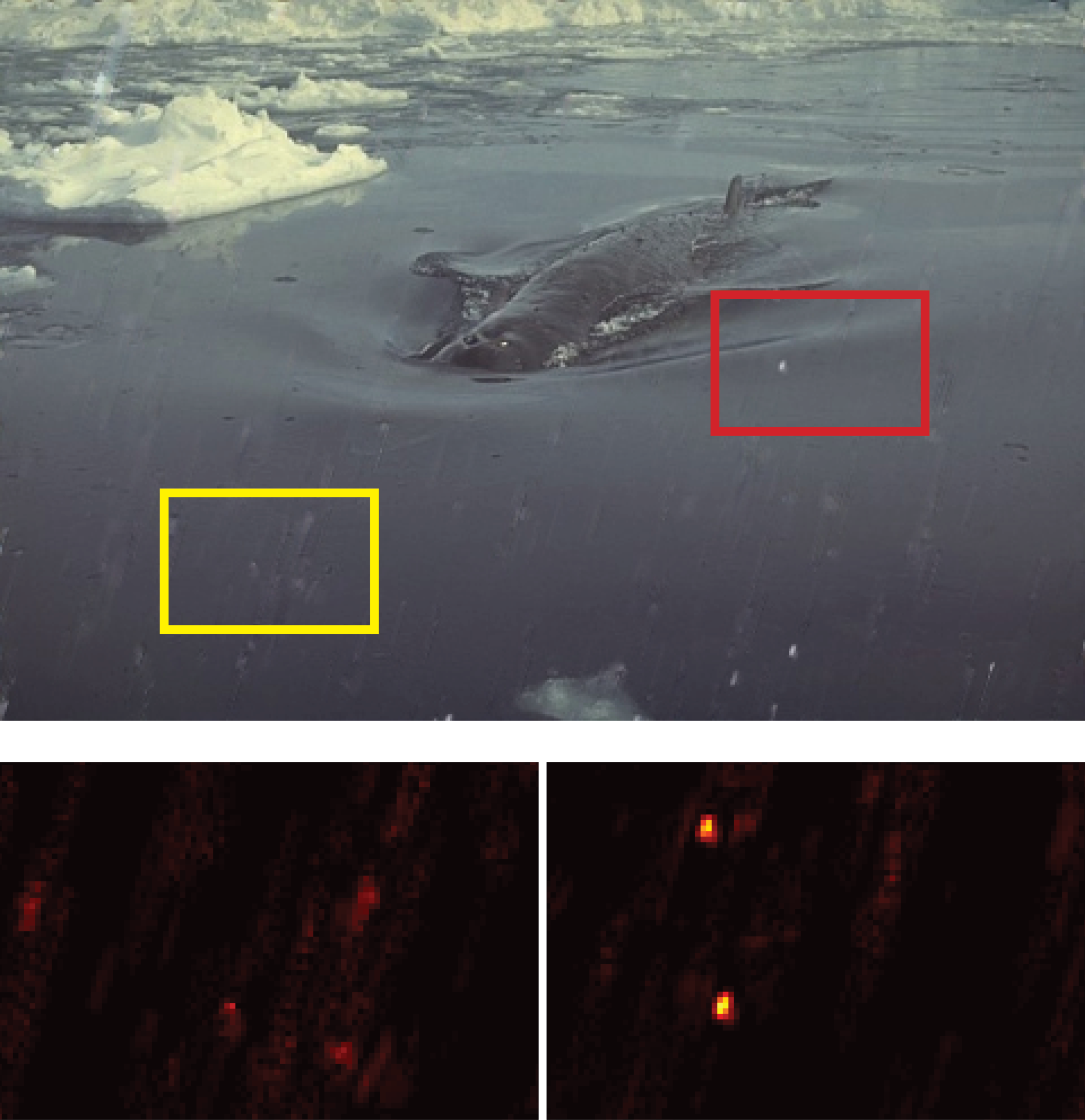} & 
        \includegraphics[width=0.19\textwidth, ]{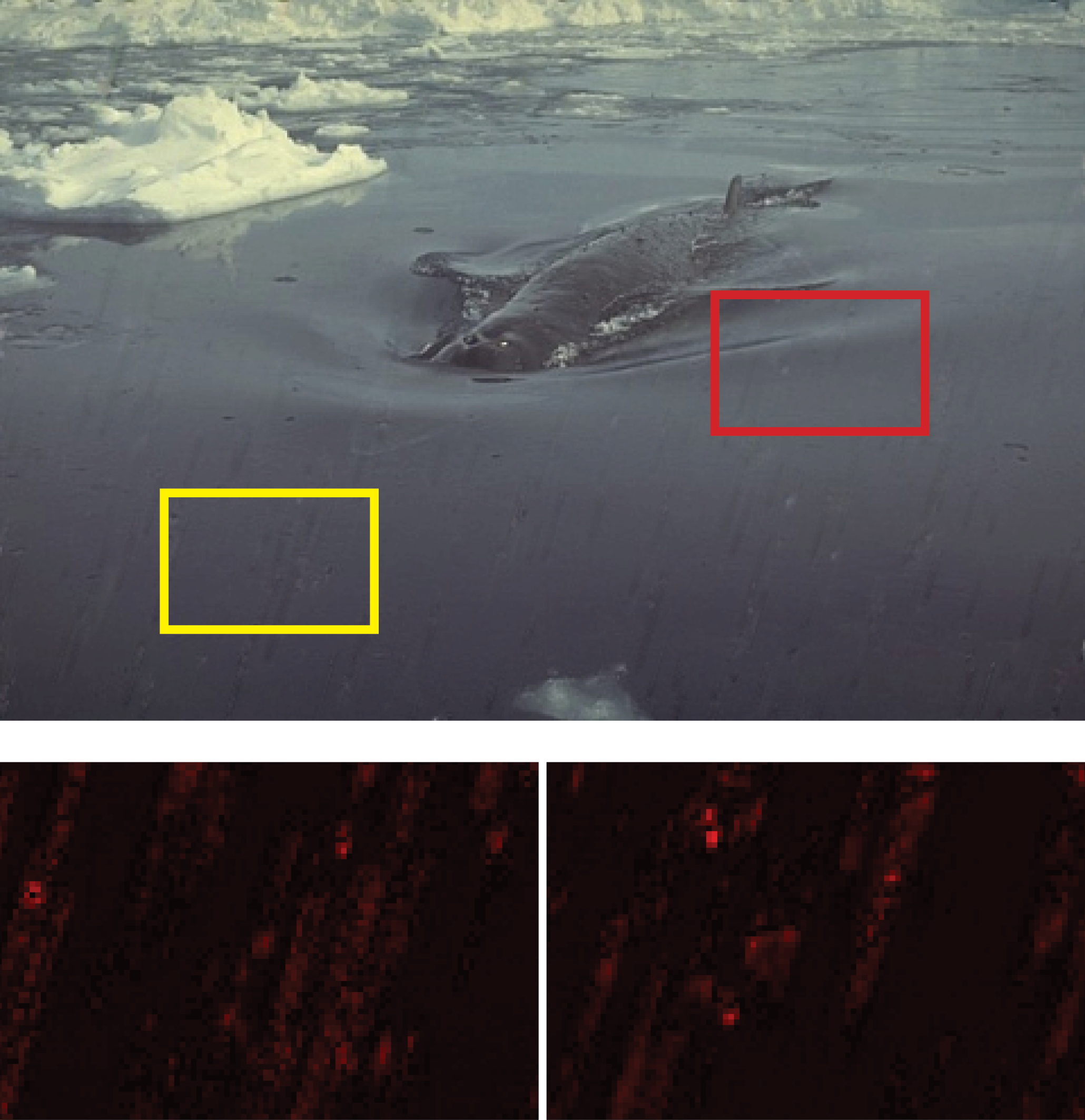} & 
        \includegraphics[width=0.19\textwidth, ]{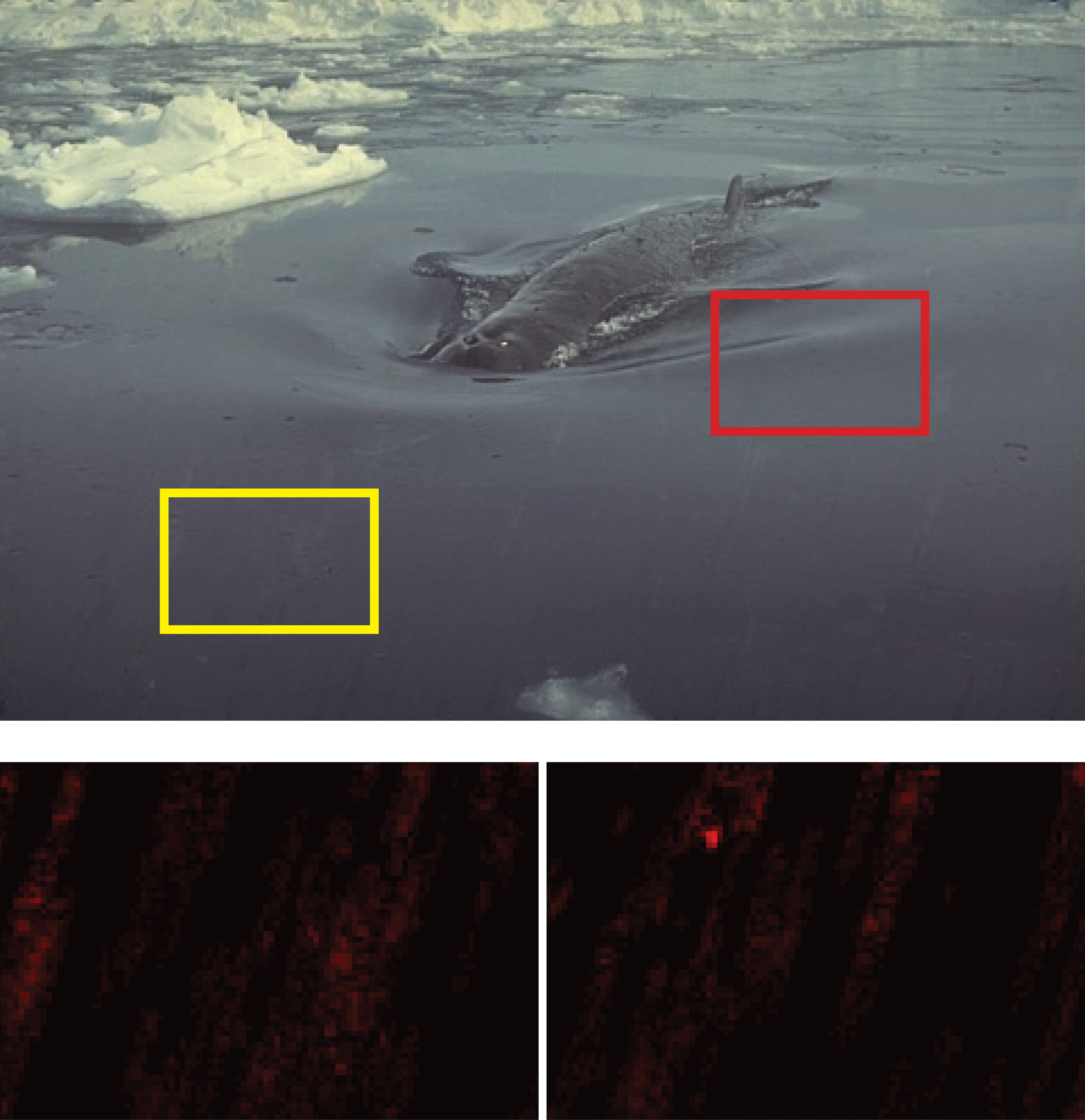} & 
        \includegraphics[width=0.19\textwidth, ]{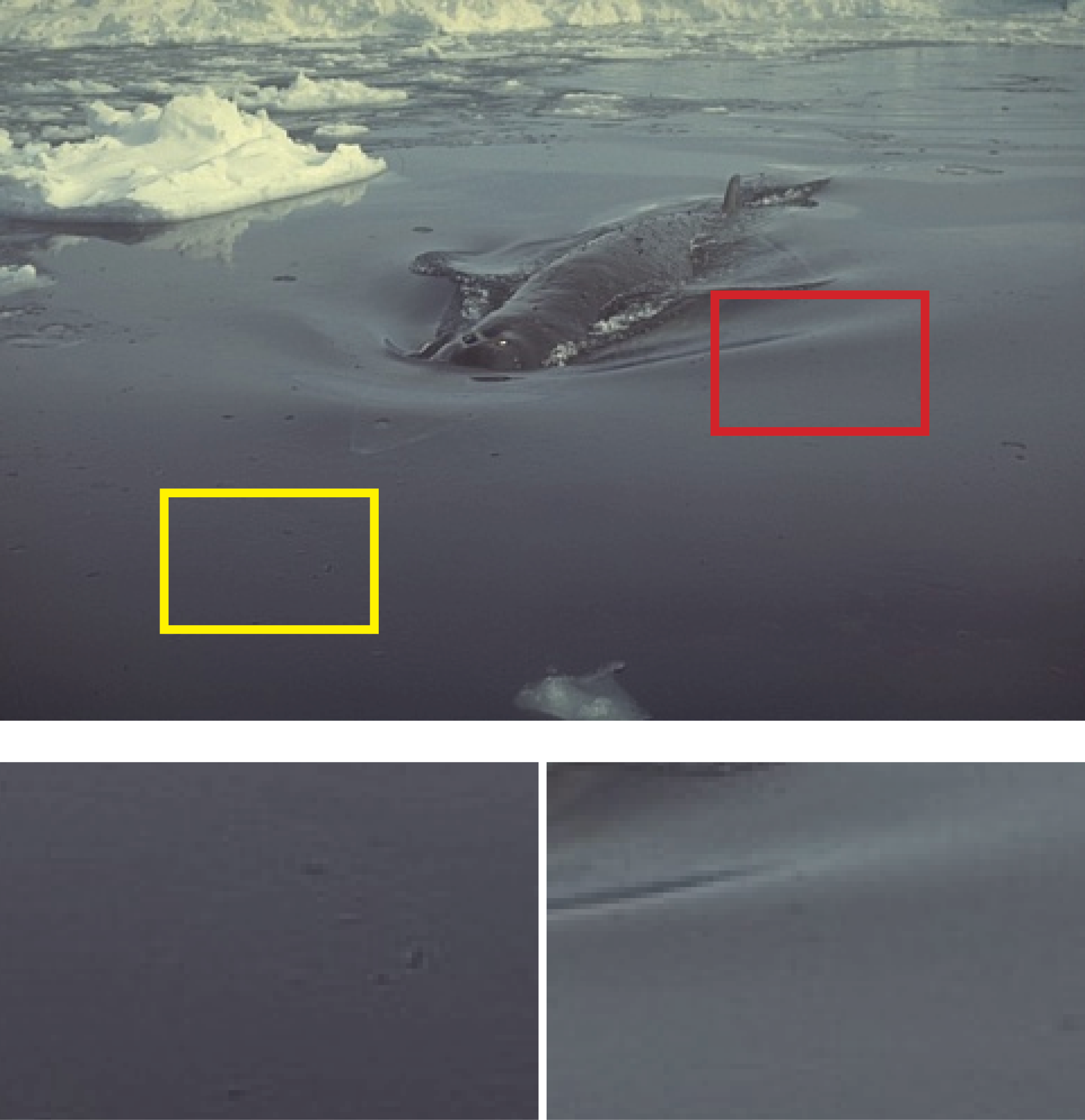}\\
    \end{tblr}
    \end{subfigure}
\hfill
    \begin{subfigure}{\textwidth}
        \begin{tblr}{
      colspec = {@{}lX[c]X[c]X[c]X[c]X[c]@{}}, colsep=0.01pt, rows={rowsep=0.5pt}, stretch = 0,
        }
        \SetCell[r=1]{l}{\rotatebox{90}{\hspace{10mm}Denoising}} &  
        \includegraphics[width=0.19\textwidth, ]{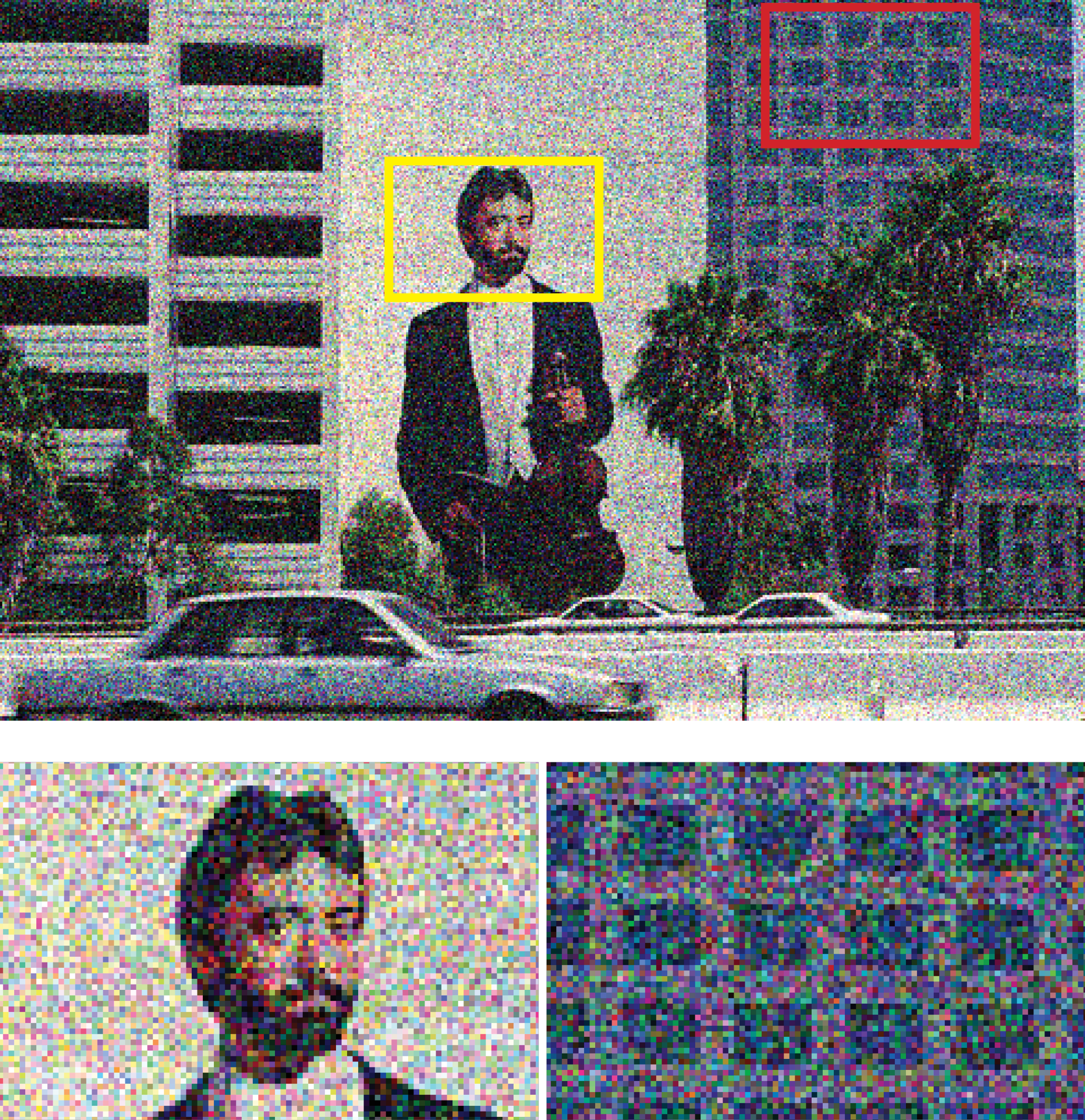} & 
        \includegraphics[width=0.19\textwidth, ]{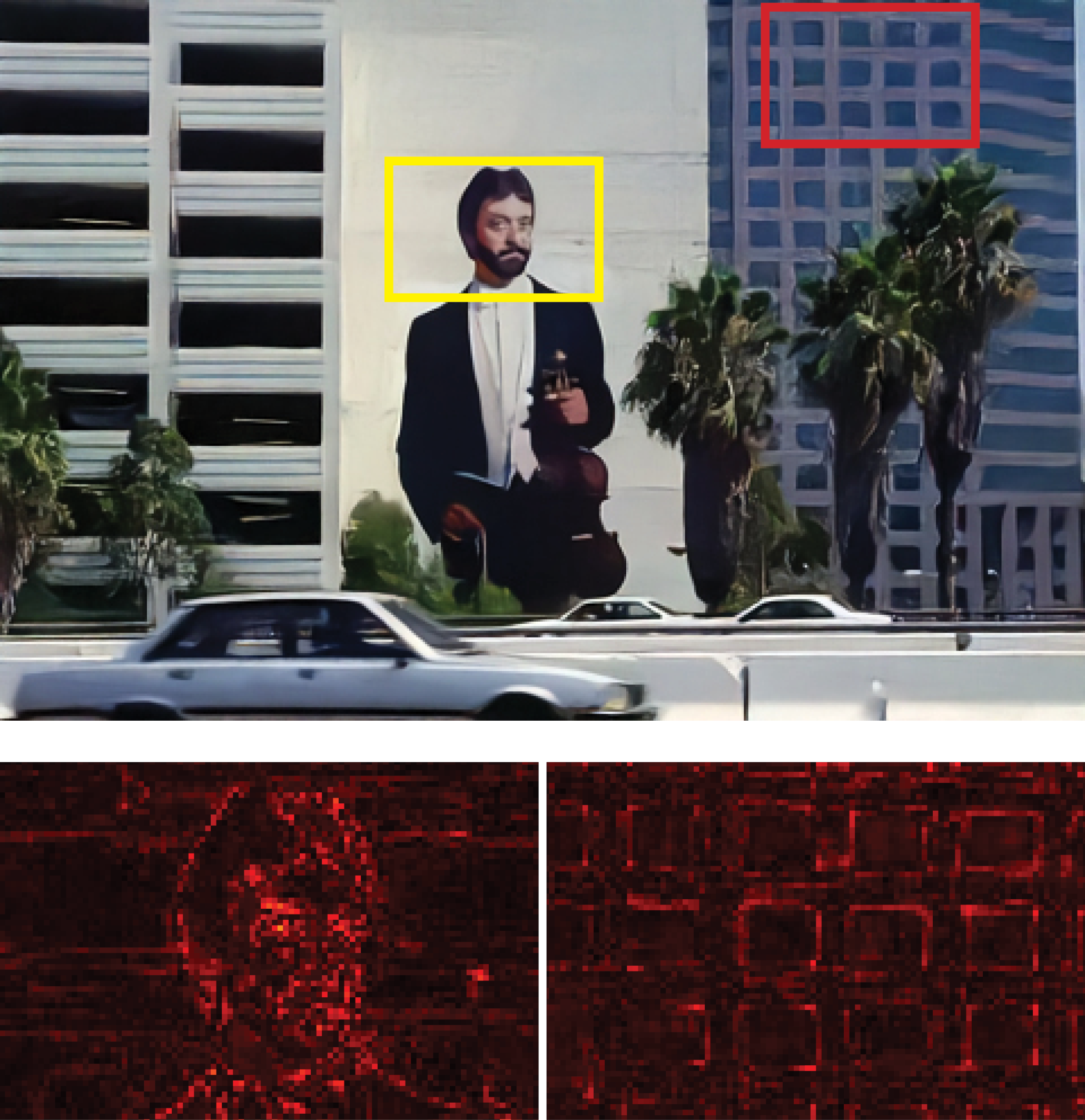} & 
        \includegraphics[width=0.19\textwidth, ]{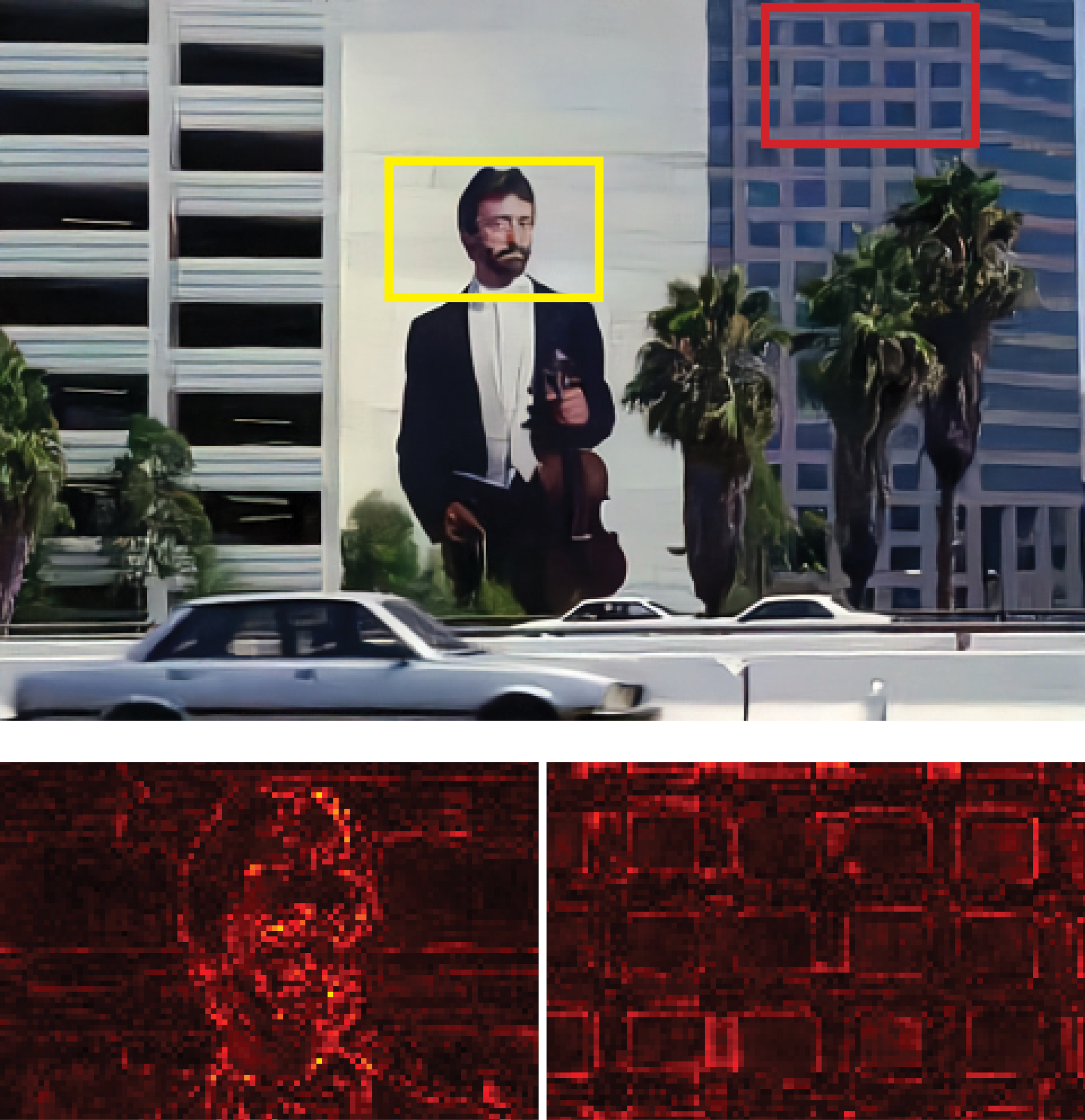} & 
        \includegraphics[width=0.19\textwidth, ]{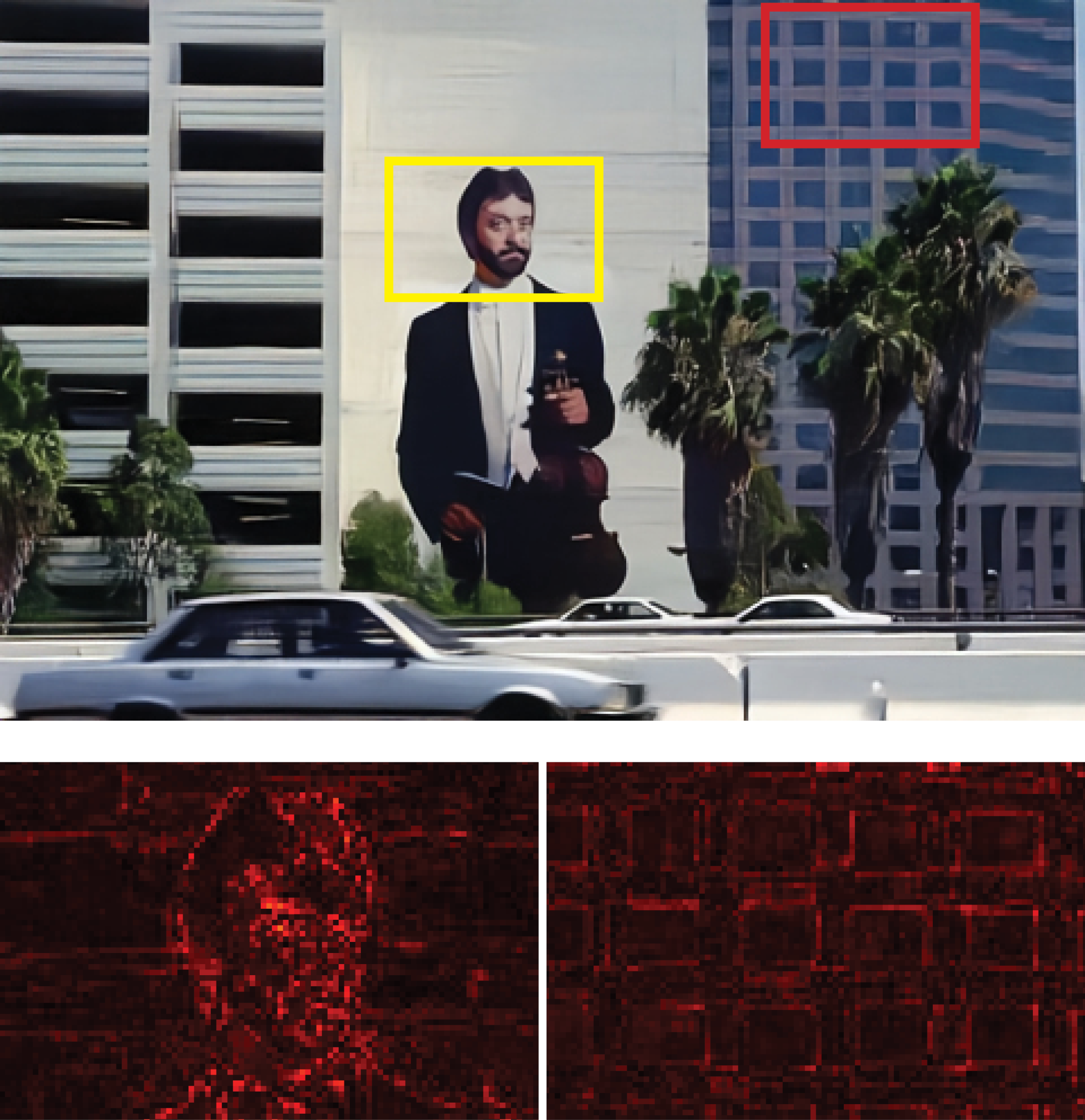} & 
        \includegraphics[width=0.19\textwidth, ]{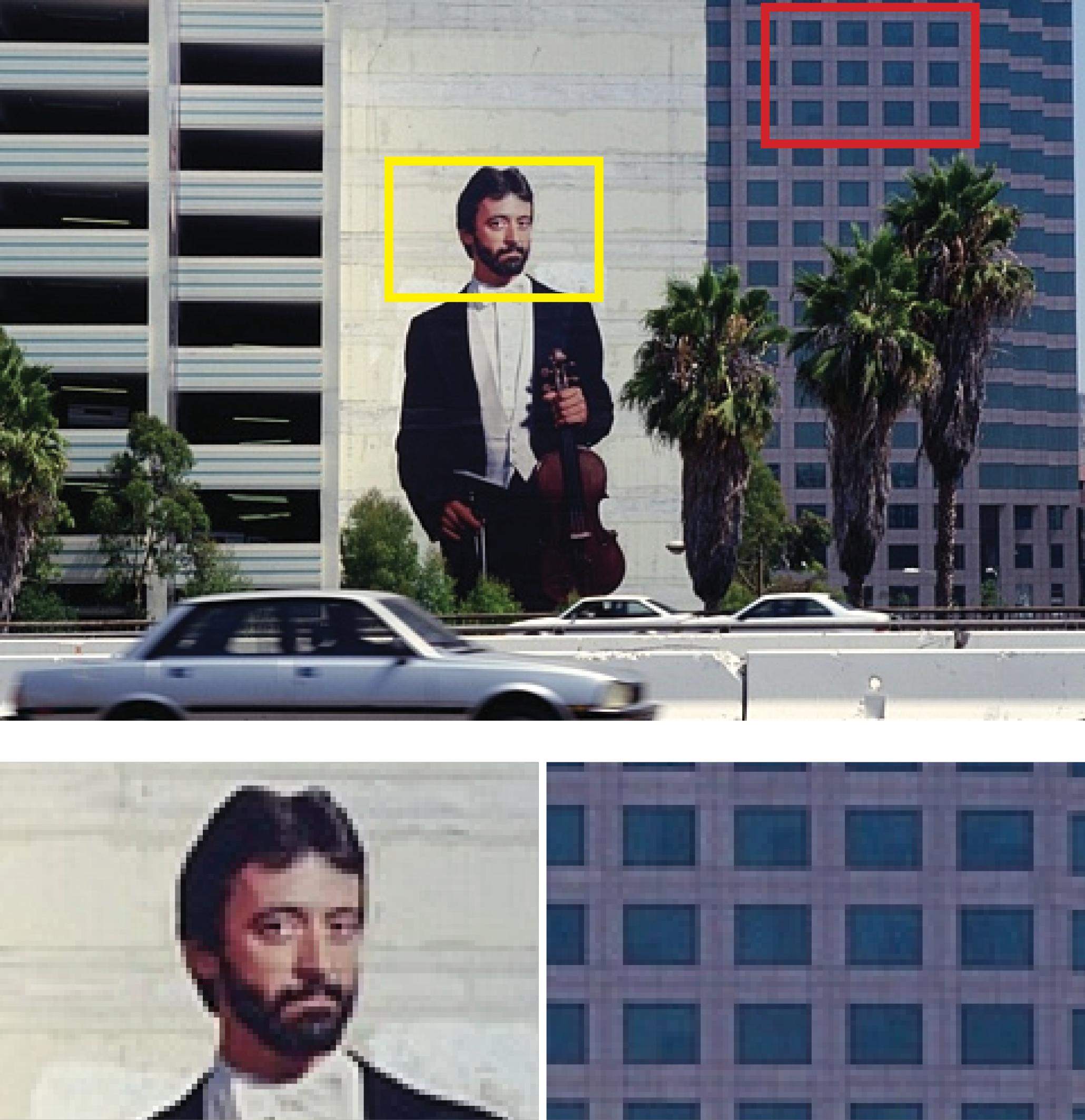}\\
        & \SetCell[c=1]{c}{{Input Crops}} &  \SetCell[c=3]{c}{{Error Maps (darker means better)}} &&& \SetCell[c=1]{c}{{GT Crops}}\\
        \end{tblr}

    \end{subfigure}
    \vspace{-2mm}
    \caption{\textit{Visual results.} We compare MoCE-IR-S to AirNet~\cite{li2022airnet}, and PromptIR~\cite{potlapalli2023promptir} in the all-in-one setting with three degradations. MoCE-IR-S effectively removes haze and rain streaks while preserving image sharpness, achieving high-quality restoration. An error heatmap is provided, with color transitioning from black to white to indicate increasing pixel-wise error.}
    \label{fig:exp:visuals}
    \vspace{-5mm}
\end{figure*}

We conduct experiments by strictly following previous works in general image restoration~\cite{potlapalli2023promptir,zhang2023ingredient} under two different settings: \textit{(i) All-in-One} and \textit{(ii) Composited degradations}. In the all-in-one setting, a unified model is trained across multiple degradation types, where we consider \textit{three} and \textit{five} distinct degradations. In the Composited degradations setting, a single model is trained to handle both isolated degradation occurrences as well as combinations of up to three different types of degradations. 

\vspace{1mm}
\myparagraphnospace{Implementation Details.}
Our MoCE-IR framework is end-to-end trainable requiring no multi-stage optimization of any components. The architecture features an asymmetrical, four-level encoder-decoder design. The encoder consists of four levels with varying numbers of transformer blocks, arranged as $[4, 6, 6, 8]$ from the top level to the lowest. The decoder includes three levels, containing $[2, 4, 4]$ transformer blocks, respectively. In each MoCE layer, we use \( n = 4 \) nested experts, where the dimensionality of the expert embedding is given by \( R = \nicefrac{C}{2^{i}}~\text{for}~i \in \{i,...,n\} \). The network’s initial embedding dimension is set to \( C = 32 \), doubling at each subsequent level while the spatial resolution is halved. We follow the training configuration of prior work~\cite{potlapalli2023promptir}, and train our models for $120$ epochs with a batch size of $32$. We optimize the $L_{1}$ loss in the RGB and Fourier domain using the Adam~\cite{kingma2017adam} optimizer ($\beta_1=0.9$, $\beta_2=0.999$) with an initial learning rate of $\num{2e-4}$ and cosine decay. During training, we utilize crops of size $128^2$ with horizontal and vertical flips as augmentations.

\vspace{1mm}
\myparagraphnospace{Datasets.}
For all-in-one, we follow existing work~\cite{li2022airnet,potlapalli2023promptir} and include following datasets: For image denoising, we combine the BSD400 \cite{arbelaez2010contour} and WED \cite{ma2016waterloo} datasets, adding Gaussian noise at levels $\sigma \in [15, 25, 50]$ to create noisy images. Testing is conducted on the BSD68 \cite{martin2001database}. For deraining, we use Rain100L \cite{yang2020learning}. The dehazing task utilises the SOTS \cite{li2018benchmarking} dataset. For deblurring and low-light enhancement, we employ the GoPro \cite{nah2017deep} and the LOL-v1 \cite{wei2018deep} dataset, respectively. To develop a unified model for all tasks, we merge these datasets in a \textit{three} (AIO-3) or \textit{five} (AIO-5) degradation setting, and train for 120 epochs and directly evaluate them across different tasks. For the \textit{composite} degradation setting, we use the CDD11 dataset~\cite{guo2024onerestore} and train our method for 200 epochs, keeping all other settings unchanged.

\subsection{Comparison to State-of-the-Art Methods}

\vspace{1mm}
\myparagraphnospace{All-in-One: Three Degradations.}
We compare our all-in-one restoration approach with specialized restoration methods, including both image-only models, such as MPRNet~\citep{zamir2021pmrnet}, AirNet~\citep{li2022airnet}, PromptIR~\citep{potlapalli2023promptir}, and Gridformer~\cite{wang2024gridformer}, and vision-language models like DA-CLIP~\cite{luo2023daclip}, InstructIR~\cite{conde2024high}, and UniProcessor~\cite{duan2024uniprocessor}. All methods were trained simultaneously on three types of degradation: haze, rain, and Gaussian noise. 
The results in \cref{tab:exp:3deg} show that our proposed model, MoCE-IR, consistently outperforms previous works, both in light and heavy model scales. In the light category, MoCE-IR achieves the highest scores across all benchmarks with an average improvement in PSNR of $1.37$~dB, surpassing other models like AirNet~\cite{li2022airnet}. In the heavy scale, MoCE-IR competes closely with UniProcessor~\cite{duan2024uniprocessor}, achieving an average PSNR of 32.73 dB with 98$\%$ fewer parameters, validating our efficiency and effectiveness.

\vspace{1mm}
\myparagraphnospace{All-in-One: Five Degradations.}
Based upon~\citep{li2022airnet,zhang2023ingredient}, we expand the three-degradation setting to include deblurring and low-light image enhancement, further validating our method’s effectiveness. As shown in \cref{tab:exp:5deg}, our approach excels by learning specialized experts with input-adaptive complexity and maximizing synergies across degradations. Our framework outperforms previous methods on the lightest model scale, with an average PSNR improvement of $4.31$ dB over AirNet~\cite{li2022airnet}. Even compared to larger networks, our lightest model surpasses the VLM-based InstructIR~\cite{conde2024high} by $0.33$ dB on average, and on a comparable model scale, our framework extends this advantage by an additional $0.50$ dB.

\vspace{1mm}
\myparagraphnospace{All-in-One: Composited Degradations.}
To simulate more realistic degradation scenarios, recent work~\cite{guo2024onerestore} has trained all-in-one restorers not only on a mix of different degradations but also on more challenging cases where multiple degradation types are combined within a single image. We extend the previous AIO settings by including rain, haze, snow, and low illumination as individual degradation types, along with various composite degradation scenarios, resulting in a total of eleven distinct restoration settings. As shown in \cref{tab:exp:cdd11}, our method outperforms previous all-in-one restorers by an average improvement of $0.64$ dB over OneRestore~\cite{guo2024onerestore}, while being among the smallest models in size. This further validates our effectiveness with adaptive modeling capacity based on input complexity.

\begin{figure}[t]
    \centering
    \scriptsize
    \fboxsep0.75pt
    \setlength\tabcolsep{5pt}
    
    \begin{tikzpicture}
\begin{groupplot}[
    group style={
        group size=2 by 1, 
        horizontal sep=0.5cm, 
        vertical sep=0cm, 
    },
    width=0.6\columnwidth,
    height=0.5\columnwidth,
    xlabel={Parameters (M)},
    ylabel={PSNR (dB)},
    xmin=0,
    xmax=50,
    grid,
    grid style=dashed,
    title style={anchor=north, at={(0.5,1.05)},font=\scriptsize},
    tick label style={font=\scriptsize},
    label style={font=\scriptsize},
    enlarge y limits=0.2,
    enlarge x limits=0.1,
    ]

\nextgroupplot[
    ylabel={PSNR (dB)},
    title={AIO-3}
]
\addplot[
    nodes near coords style={font=\scriptsize, anchor=south west, xshift=-2pt},
    scatter/classes={
            a={color=tabblue, mark size=1.7},
            b={color=tabblue!50, mark size=1.7},
            c={color=taborange!50, mark size=1.7},
            d={color=taborange, mark size=1.7}
            },
    scatter, mark=*, only marks, 
    scatter src=explicit symbolic,
    nodes near coords*={\Label},
    visualization depends on={value \thisrow{label} \as \Label} 
    ] table [meta=class] {
    x y class label
    2 29.98 b DL
    9 31.20 b AirNet
    16 30.17 b MPRNet
    36 32.06 b PromptIR
    33 32.49 b Art
    };

\addplot[
    nodes near coords style={font=\scriptsize, anchor=south west, xshift=-2pt},
    scatter/classes={
            a={color=tabblue, mark size=1.7},
            b={color=tabblue!50, mark size=1.7},
            c={color=taborange!50, mark size=1.7},
            d={color=taborange, mark size=1.7}
            },
    scatter, mark=*, only marks, 
    scatter src=explicit symbolic,sharp plot, tabblue, line width=0.75pt,
    nodes near coords*={\Label},
    visualization depends on={value \thisrow{label} \as \Label} 
    ] table [meta=class] {
    x y class label
    11 32.57 a \textcolor{white}{ours}
    25 32.73 a Ours
    };

\nextgroupplot[
    ylabel={},
    title={AIO-5}
]
\addplot[
    nodes near coords style={font=\scriptsize, anchor=north west, xshift=-2pt},
    scatter/classes={
            a={color=tabblue, mark size=1.7},
            b={color=tabblue!50, mark size=1.7},
            c={color=taborange!50, mark size=1.7},
            d={color=taborange, mark size=1.7}
            },
    scatter, mark=*, only marks, 
    scatter src=explicit symbolic,
    nodes near coords*={\Label},
    visualization depends on={value \thisrow{label} \as \Label} 
    ] table [meta=class] {
    x y class label
    1 25.09 c TAPE
    17 29.33 c InstructIR
    15 28.34 c IDR
    };

\addplot[
    nodes near coords style={font=\scriptsize, anchor=south west, xshift=-2pt},
    scatter/classes={
            a={color=tabblue, mark size=1.7},
            b={color=tabblue!50, mark size=1.7},
            c={color=taborange!50, mark size=1.7},
            d={color=taborange, mark size=1.7}
            },
    scatter, mark=*, only marks, 
    scatter src=explicit symbolic,
    nodes near coords*={\Label},
    visualization depends on={value \thisrow{label} \as \Label} 
    ] table [meta=class] {
    x y class label
    9 25.49 c AirNet
    34 29.33 c Gridformer
    };

\addplot[
    nodes near coords style={font=\scriptsize, anchor=south west, xshift=-2pt},
    scatter/classes={
            a={color=tabblue, mark size=1.7},
            b={color=tabblue!50, mark size=1.7},
            c={color=taborange!50, mark size=1.7},
            d={color=taborange, mark size=1.7}
            },
    scatter, mark=*, only marks, 
    scatter src=explicit symbolic,sharp plot, taborange, line width=0.75pt,
    nodes near coords*={\Label},
    visualization depends on={value \thisrow{label} \as \Label} 
    ] table [meta=class] {
    x y class label
    11 29.88 d  \textcolor{white}{3M}
    25 30.58 d Ours
    };

\end{groupplot}
\end{tikzpicture}
    \vspace{-7mm}
    \captionof{figure}{\textit{Complexity-efficiency tradeoff.} Visualization of PSNR and parameter counts of proposed method compared to prior work. Proposed MoCE-IR surpasses prior methods, achieving SoTA results in all-in-one image restoration with enhanced efficiency.}
    \label{fig:exp:tradeoff}
    \vspace{-2mm}
    \captionof{table}{\textit{Computational demands.} GFLOPS are computed on an input image of size $224 \times 224$ using a NVIDIA RTX 4090 GPU.}
    \label{tab:exp:computational_demands}
    \begin{tabularx}{\columnwidth}{X*{4}{c}}
    \toprule
    Method & Params. & Memory & FLOPS & Runtime\\
    \midrule
    AirNet~\cite{li2022airnet} & \textbf{8.93M} & 4829M  & 238G & 42.17~±~0.23\\
    PromptIR~\cite{potlapalli2023promptir}&  35.59M & 9830M & 132G & 41.28~±~0.43 \\
    IDR~\cite{zhang2023ingredient}& 15.34M & 4905M  & 98G & - \\
    \rowcolor{gray!10}
    MoCE-IR~(\textit{ours})& 25.35M & 5887M & 80.59~±~5.21G & 23.36~±~2.47\\ 
    \rowcolor{gray!10}
    MoCE-IR-S~(\textit{ours})& 11.47M & \textbf{4228M} & \textbf{36.93~±~2.32G} & \textbf{22.15}~±~2.59\\ 
    \bottomrule
    \end{tabularx}
    \vspace{-5mm}
\end{figure}

\vspace{1mm}
\myparagraphnospace{Visual Results.}
We show visual results obtained under the three degradation settings in \cref{fig:exp:visuals}. 
In certain demanding hazy scenarios, both AirNet~\citep{li2022airnet} and PromptIR~\citep{potlapalli2023promptir} reveal constraints in completely eradicating haziness, resulting in noticeable color intensity discrepancies, whereas our approach ensures precise color reconstruction. Furthermore, in challenging rainy scenes, these popular approaches continue to exhibit notable remnants of rain, which our method adeptly eliminates, distinguishing itself from other approaches. Additionally, our method also generates clear and sharp denoised outputs. We further provide the error maps for visual differences where our method contains less noticeable errors. Coupled with the quantitative comparisons, these findings underscore the effectiveness of our method. \textit{Please refer to the supplementary for more visuals.}

\begin{table}[t]
    \centering
    \scriptsize
    \fboxsep0.75pt
    \setlength\tabcolsep{2pt}
    \caption{\textit{Architecture analysis.} 
    (a) We investigate scaling the expert embedding  \textcolor{myred}{$r$} and the bias normalization for MoCE-IR-S. (b) We compare MoCE to several baselines~\cite{riquelme2021scaling,yang2024amir,luo2023wm}, demonstrating that our complexity experts consistently outperform them. Notably, complexity-biased routing further boosts performance, surpassing standard load balancing (LB)~\cite{riquelme2021scaling}.
    PSNR (dB, $\uparrow$) and \colorbox{clblue!50}{SSIM ($\uparrow$)} are reported on full RGB images.}
    \vspace{-3mm}
    \label{tab:exp:arch_analysis}
\begin{subtable}[t]{\columnwidth}
    \subcaption{\textit{Ablation on the expert scaling.}}
    \label{tab:exp:expert_design}
    \setlength\tabcolsep{12pt}
           \begin{tabularx}{\columnwidth}{*{4}{c}}
            \toprule
            Embedding \textcolor{myred}{$r$} scaling  & Bias normalization & \multicolumn{2}{c}{Avg. AIO-3}\\
            \midrule
                  Nested & $p_{\text{Min}}$ & 32.44	&\cc{.913} \\
                  Exponential & $p_{\text{Max}}$ & 32.47 &\cc{.914} \\
                  Nested & $p_{\text{Max}}$ & \textbf{32.57} &\cc{\textbf{.916}} \\
            \bottomrule
            \end{tabularx}   
    \end{subtable}%
\hfill
\begin{subtable}[t]{\columnwidth}
    \setlength\tabcolsep{1.5pt}
    \subcaption{\textit{Ablation on the complexity-aware routing.}}
    \label{tab:exp:complexity_bias}
       \begin{tabularx}{\columnwidth}{X*{8}{c}}
       \toprule
        Method & \multicolumn{2}{c}{SOTS} & \multicolumn{2}{c}{Rain100L} & \multicolumn{2}{c}{$\text{BSD68}_{\text{15-50}}$} & \multicolumn{2}{c}{Avg.}\\
       \midrule
       MoCE + Complexity Bias (ours) &\textbf{ 30.94} & \cc{\textbf{.979}} & 38.22 & \cc{\textbf{.983}} & \textbf{31.22} & \cc{\textbf{.873}} & \textbf{32.57} & \cc{\textbf{.916}} \\
       \midrule
       Baseline (MoE + LB~[\textcolor{tabblue}{45}]) & 30.28 & \cc{.977} & 37.77 & \cc{.981} & 31.14 & \cc{.871} & 32.30 & \cc{.914} \\
       \ w/ MoCE + LB~[\textcolor{tabblue}{45}] & 30.13 & \cc{.978} & \textbf{38.24} & \cc{.983} & 31.19 & \cc{.872} & 32.39 & \cc{.915} \\
       \ w/ AMIR~\cite{yang2024amir} & 29.97 & \cc{.975} & 37.90 & \cc{.981} & 31.17 & \cc{.872} & 32.28 & \cc{.914} \\
       \ w/ WM-MoE~\cite{luo2023wm} & 30.10 & \cc{.976} & 36.72	& \cc{.976}& 31.05 & \cc{.869} & 31.99 & \cc{.911} \\
       \bottomrule
       \end{tabularx}
\end{subtable}%
\vspace{-5mm}
\end{table}

\subsection{Ablation Studies}
We conduct detailed studies on the components within our method, highlighting the emergence of task-discriminative learners.  All experiments are conducted in the all-in-one setting with three degradations. \textit{Please refer to the supplementary for additional ablations.}

\vspace{1mm}
\myparagraphnospace{Complexity-Efficiency Trade-off.}
Our MoCE-IR framework addresses the critical challenge of scalability in vision tasks by optimizing both reconstruction quality and computational efficiency. We visualize in \cref{fig:exp:tradeoff} the progressive improvements in reconstruction fidelity of our framework on two complexity scales. Furthermore, \cref{tab:exp:computational_demands} validates that our approach significantly outperforms SoTA methods in terms of memory utilization.
Unlike traditional MoE architectures with fixed computational demands, our complexity experts enable dynamic FLOPS allocation. For instance, our lightweight model averages $36.9$ GFLOPS, while the heavyweight model averages $80.59$ GFLOPS.
This adaptive computation, combined with lower memory consumption and parameter counts, gives MoCE-IR a substantial advantage over contemporary methods.

\begin{figure*}[t]
        \centering
        \begin{tikzpicture}[baseline]
  \begin{groupplot}[
    group style={group size=4 by 1, vertical sep=0.4cm, horizontal sep=0.5cm},
    width=0.30\textwidth, height=0.2\textwidth, 
    ytick={1,2,3,4}, ymin=1,ymax=4,
    yticklabels={1,2,3,4},
    xtick pos=left, ytick pos=left, xlabel near ticks, ylabel near ticks, xtick=data, enlargelimits=0.05,
    tick label style={font=\scriptsize}, title style={font=\scriptsize}, label style={font=\scriptsize}, tick align=inside,
    xlabel={Decoder Layers},
    ylabel style={anchor=center, at={(-0.15,0.5)},},
    ]

 \nextgroupplot[
    ylabel={Experts}, 
    extra y ticks={1, 4}, 
    extra y tick labels={Low, High}, 
    extra y tick style={ticklabel style={anchor=north, font=\scriptsize, rotate=90, yshift=22pt}}, 
    title={(a) Load balancing~\cite{riquelme2021scaling}}, title style={anchor=north, at={(0.5,1.1)},},
    ]
    \addplot [tabblue, thick] coordinates { 
        (1, 3) (2, 2) (3, 1) (4, 2) (5, 1) (6, 3) (7, 2) (8, 4) (9, 3) (10, 3)
    };
    \addplot[name path=dehaze_lower, white] coordinates { 
        (1, 1.8922) (2, 0.8892) (3, -0.1353) (4, 0.9128) (5, -0.1062) (6, 1.8876) (7, 0.8695) (8, 2.8802) (9, 1.8784) (10, 1.8809)
    };
    \addplot[name path=dehaze_upper, white] coordinates { 
        (1, 4.1078) (2, 3.1108) (3, 2.1353) (4, 3.0872) (5, 2.1062) (6, 4.1124) (7, 3.1305) (8, 5.1198) (9, 4.1216) (10, 4.1191)
    };
    \addplot[fill=tabblue, fill opacity=0.2] fill between[of=dehaze_lower and dehaze_upper];
    
    \addplot [taborange, thick] coordinates { 
        (1, 2) (2, 3) (3, 1) (4, 1) (5, 2) (6, 3) (7, 4) (8, 3) (9, 1) (10, 3)
    };
    \addplot[name path=derain_lower, white] coordinates { 
        (1, 0.9449) (2, 1.9469) (3, -0.1840) (4, -0.1176) (5, 0.9022) (6, 1.9488) (7, 2.8649) (8, 1.8938) (9, -0.1531) (10, 1.9103)
    };
    \addplot[name path=derain_upper, white] coordinates { 
        (1, 3.0550) (2, 4.0531) (3, 2.1840) (4, 2.1176) (5, 3.0978) (6, 4.0512) (7, 5.1351) (8, 4.1062) (9, 2.1531) (10, 4.0897)
    };
    \addplot[fill=taborange, fill opacity=0.2] fill between[of=derain_lower and derain_upper];
    
    \addplot [tabgreen, thick] coordinates { 
        (1, 2) (2, 3) (3, 1) (4, 3) (5, 2) (6, 2) (7, 2) (8, 3) (9, 4) (10, 4)
    };
    \addplot[name path=denoise_15_lower, white] coordinates { 
        (1, 0.9532) (2, 1.9271) (3, -0.1833) (4, 1.9225) (5, 0.9524) (6, 1.0075) (7, 0.9000) (8, 1.9395) (9, 2.8767) (10, 2.8620)
    };
    \addplot[name path=denoise_15_upper, white] coordinates { 
        (1, 3.0468) (2, 4.0729) (3, 2.1833) (4, 4.0775) (5, 3.0476) (6, 2.9925) (7, 3.1000) (8, 4.0605) (9, 5.1233) (10, 5.1380)
    };
    \addplot[fill=tabgreen, fill opacity=0.2] fill between[of=denoise_15_lower and denoise_15_upper];

 \nextgroupplot[
    ylabel=\empty, 
    title={(b) \textit{Ours} on SOTS}, title style={anchor=north, at={(0.5,1.1)},},
    ]
    \addplot [tabblue, thick] coordinates { 
        (1, 4) (2, 2) (3, 4) (4, 2) (5, 4) (6, 2) (7, 2) (8, 2) (9, 1) (10, 1)
    };
    \addplot[name path=dehaze_lower, white] coordinates { 
        (1, 2.8659) (2, 1.2010) (3, 2.8225) (4, 1.2000) (5, 2.8414) (6, 1.1950) (7, 1.2114) (8, 1.2022) (9, 0.1832) (10, 0.1912)
    };
    \addplot[name path=dehaze_upper, white] coordinates { 
        (1, 5.1340) (2, 2.7989) (3, 5.1774) (4, 2.7999) (5, 5.1585) (6, 2.8049) (7, 2.7885) (8, 2.7977) (9, 1.8167) (10, 1.8087)
    };
    \addplot[fill=tabblue, fill opacity=0.2] fill between[of=dehaze_lower and dehaze_upper];

 \nextgroupplot[
    ylabel=\empty, 
    title={(c) \textit{Ours} on Rain100L}, title style={anchor=north, at={(0.5,1.1)},},
    ]
    \addplot [taborange, thick] coordinates { 
        (1, 2) (2, 2) (3, 1) (4, 1) (5, 2) (6, 2) (7, 1) (8, 3) (9, 1) (10, 1)
    };
    \addplot[name path=derain_lower, white] coordinates { 
        (1, 1.2324) (2, 1.2482) (3, 0.1512) (4, 0.1722) (5, 1.1875) (6, 1.2148) (7, 0.1968) (8, 2.1574) (9, 0.1763) (10, 0.1820)
    };
    \addplot[name path=derain_upper, white] coordinates { 
        (1, 2.7675) (2, 2.7517) (3, 1.8487) (4, 1.8277) (5, 2.8124) (6, 2.7851) (7, 1.8031) (8, 3.8425) (9, 1.8236) (10, 1.8179)
    };
    \addplot[fill=taborange, fill opacity=0.2] fill between[of=derain_lower and derain_upper];

 \nextgroupplot[
    ylabel=\empty,
    title={(d) \textit{Ours} on $\text{BSD68}_{\sigma=15}$}, title style={anchor=north, at={(0.5,1.1)},},
    ]

    \addplot [tabgreen, thick] coordinates { 
        (1, 2) (2, 4) (3, 1) (4, 4) (5, 3) (6, 2) (7, 4) (8, 4) (9, 4) (10, 4)
    };
    \addplot[name path=denoise15_lower, white] coordinates { 
        (1, 1.2146) (2, 2.9359) (3, 0.1854) (4, 2.9094) (5, 1.1898) (6, 1.2439) (7, 2.9166) (8, 2.9303) (9, 2.9117) (10, 2.9518)
    };
    \addplot[name path=denoise15_upper, white] coordinates { 
        (1, 2.7853) (2, 5.0640) (3, 1.8145) (4, 5.0905) (5, 4.8101) (6, 2.7560) (7, 5.0833) (8, 5.0696) (9, 5.0882) (10, 5.0481)
    };
    \addplot[fill=tabgreen, fill opacity=0.2] fill between[of=denoise15_lower and denoise15_upper];

  \end{groupplot}
    \end{tikzpicture}
        \vspace{-6mm}
        \caption{\textit{Routing visualization for the AIO-3 setting.} 
        (a) While load balancing~\cite{riquelme2021scaling} ensures uniform expert utilization, it neglects shared task dependencies and task-specific characteristics, limiting restoration quality.
        (b)-(d) Complexity-aware routing fosters task discrimination by directing complex degradations to experts with broader contextual understanding and vice versa.
        This allows some experts to generalize across tasks while others specialize, enhancing adaptability.
        We visualize the average decisions made by each router for dehazing, deraining and denoising. The y-axis indicates increasing expert complexity.
        }
        \vspace{-5mm}
        \label{fig:exp:routing_aio3}
\end{figure*}

\vspace{1mm}
\myparagraphnospace{Expert Scaling.} As shown in \cref{tab:exp:expert_design}, we systematically evaluate different scaling strategies regarding the embedding dimensionality \textcolor{myred}{$r$} of each complexity expert. Our investigation compares exponential and nested parameter scaling, where the window partition is always fixed and follows a \(2^{(2 + i)}\) progression. Further, we investigate the bias normalization, whether normalizing to the lowest or highest expert capacity. Overall, the nested parameter scaling emerges as the optimal choice, providing sufficient expert channel diversity while maintaining reasonable computational costs. Normalizing router weights by the highest expert capacity, rather than the minimum, proves crucial for performance as minimum capacity normalization forces router weights of higher capacity experts to escalate uncontrollably, thus excluding them from the gating mechanism.

\vspace{1mm}
\myparagraphnospace{Impact of Complexity-aware Routing.} Building on the findings from \cref{tab:exp:expert_design}, we adopt nested expert scaling and evaluate our complexity-aware routing strategy against conventional load balancing~\cite{riquelme2021scaling}, which enforces uniform expert usage. 
As shown in \cref{tab:exp:complexity_bias}, our MoCE-IR framework achieves superior performance by preferentially routing to lower-complexity experts, activating higher-complexity experts only when their added capacity provides clear benefits. 
This efficient model utilization not only boosts image restoration performance, consistently surpassing related adaptive restoration methods~\cite{yang2024amir,luo2023wm}, but also enhances out-of-distribution generalization as MoCE-IR effectively handles degradations that with  significant deviation from the training data; \textit{see the supplementary for more evaluations}.

\begin{table}[t]
\centering
\scriptsize
\fboxsep0.75pt
\setlength\tabcolsep{3pt}
\caption{\textit{Expert generalization.} We report PSNR for MoCE-IR-S with expert assignments fixed post-training. $\mathcal{E}$ denotes expert.
}
\label{tab:exp:moe_generalization}
\vspace{-3mm}
   \begin{tabularx}{\columnwidth}{llC*{5}{c}}
   \toprule
     \multirow{2}{*}{Degradation} & \multirow{2}{*}{Routing} & Learned & \multicolumn{4}{c}{{Manual Choice}}\\
     &  & Choice & $\mathcal{E}_{1}$ & $\mathcal{E}_{2}$ & $\mathcal{E}_{3}$ & $\mathcal{E}_{4}$ \\
   \midrule
       Rain & \multirow{2}{*}{Load Balance~[\textcolor{tabblue}{45}]} & Not & 28.27 & 28.24 & 28.25 & 28.27 \\
       $\text{Noise}$ & & Applicable & 33.33 & 33.37 & 33.37 & 33.40 \\
   \midrule
   Rain & \multirow{2}{*}{Complexity Bias (ours)} & $\mathcal{E}_{1}$ & \textbf{30.45} & 30.21 & 29.73 & 24.65\\
   Noise & & $\mathcal{E}_{4}$& 33.92 & 33.93 & 33.93 & \textbf{34.00}  \\
   \bottomrule
\end{tabularx}
\vspace{-5mm}
\end{table}

\vspace{1mm}
\myparagraphnospace{Routing Analysis.} \cref{fig:exp:routing_aio3} illustrates the average expert selection per layer for different degradations: haze, rain, and noise. Higher-complexity experts are favored for tasks requiring broader context, like haze and noise, while lighter experts are chosen for rain streak removal. Notably, some experts handle multiple tasks, while others remain task-specific. Our complexity-conditioned routing promotes task-discriminative learning by directing inputs to relevant experts, preventing the collapse into a single-expert dependency, ensuring efficient and diverse operation.

The baseline MoE~\cite{riquelme2021scaling} approach, lacking complexity awareness and scaled experts, uniformly distributes degraded samples across all experts. Our results show that this uniform allocation is suboptimal for all-in-one restoration, as it fails to leverage task-specific commonalities and distinctions effectively. 
\cref{tab:exp:moe_generalization} highlights the limitations of standard MoEs~\cite{riquelme2021scaling}, where fixed post-training expert assignments result in uniform, task-agnostic performance, offering only moderate cross-task generalization. In contrast, MoCEs effectively generalize across degradations while exhibiting task-discriminative behavior, routing inputs to the most suitable experts for optimal performance.

\section{Conclusion, Limitations and Discussion}
\label{sec:conclusion}
This paper presents MoCE-IR, an efficient all-in-one image restoration model leveraging complexity experts, specialized units with varying computational demands and receptive fields. By aligning degradation tasks with suitable resources, MoCE-IR overcomes limitations of prior methods. Its built-in bias toward simpler paths enhances task-discriminative learning, enabling swift inference by bypassing irrelevant experts without sacrificing restoration quality.

Extensive experiments across diverse degradation settings, including more demanding scenarios with multiple composited degradations, demonstrate our computational efficiency through selective expert utilization and consistent improvement over state-of-the-art methods.

While MoCE-IR shows substantial improvements, opportunities remain. The current image-level routing imposes scalability constraints. Future work could explore Soft-MoE~\cite{puigcerver2024softmoe} architectures for fine-grained spatial routing or early-exiting approaches~\cite{raposo2024mod} to further enhance processing speeds and input adaptability. Extending to synthetic-to-real adaptation, especially for video, represents a crucial next step. Performance optimization through mixed-precision computation across experts could further enhance speed and efficiency for resource-limited deployments.

\vspace{1mm}
\myparagraphnospace{Acknowledgements.}
The authors sincerely thank the reviewers and all members of the program committee for their tremendous efforts and incisive feedback. 
This research was supported by the Alexander von Humboldt Foundation, the Ministry of Education and Science of Bulgaria (support for INSAIT, under the Bulgarian National Roadmap for Research Infrastructure), the Shanghai Municipal Science and Technology Major Project (2021SHZDZX0102), and the Fundamental Research Funds for the Central Universities.

{
    \small
    \bibliographystyle{ieeenat_fullname}
    \bibliography{main}

\begin{thebibliography}{72}
\providecommand{\natexlab}[1]{#1}
\providecommand{\url}[1]{\texttt{#1}}
\expandafter\ifx\csname urlstyle\endcsname\relax
  \providecommand{\doi}[1]{doi: #1}\else
  \providecommand{\doi}{doi: \begingroup \urlstyle{rm}\Url}\fi

\bibitem[Ai et~al.(2024)Ai, Huang, Zhou, Wang, and He]{ai2024multimodal}
Yuang Ai, Huaibo Huang, Xiaoqiang Zhou, Jiexiang Wang, and Ran He.
\newblock Multimodal prompt perceiver: Empower adaptiveness generalizability and fidelity for all-in-one image restoration.
\newblock In \emph{CVPR}, 2024.

\bibitem[Arbelaez et~al.(2010)Arbelaez, Maire, Fowlkes, and Malik]{arbelaez2010contour}
Pablo Arbelaez, Michael Maire, Charless Fowlkes, and Jitendra Malik.
\newblock Contour detection and hierarchical image segmentation.
\newblock \emph{TPAMI}, 33\penalty0 (5):\penalty0 898--916, 2010.

\bibitem[Bengio et~al.(2013)Bengio, L{\'e}onard, and Courville]{bengio2013estimating}
Yoshua Bengio, Nicholas L{\'e}onard, and Aaron Courville.
\newblock Estimating or propagating gradients through stochastic neurons for conditional computation.
\newblock \emph{arXiv preprint arXiv:1308.3432}, 2013.

\bibitem[Berman and Avidan(2016)]{berman2016non}
Dana Berman and Shai Avidan.
\newblock Non-local image dehazing.
\newblock In \emph{CVPR}, 2016.

\bibitem[Cai et~al.(2016)Cai, Xu, Jia, Qing, and Tao]{cai2016dehazenet}
Bolun Cai, Xiangmin Xu, Kui Jia, Chunmei Qing, and Dacheng Tao.
\newblock Dehazenet: An end-to-end system for single image haze removal.
\newblock \emph{TIP}, 25\penalty0 (11):\penalty0 5187--5198, 2016.

\bibitem[Chen et~al.(2022{\natexlab{a}})Chen, Chu, Zhang, and Sun]{chen2022simple}
Liangyu Chen, Xiaojie Chu, Xiangyu Zhang, and Jian Sun.
\newblock Simple baselines for image restoration.
\newblock In \emph{ECCV}, 2022{\natexlab{a}}.

\bibitem[Chen and Pei(2023)]{chen2023alwayscleardays}
Yu-Wei Chen and Soo-Chang Pei.
\newblock Always clear days: Degradation type and severity aware all-in-one adverse weather removal.
\newblock \emph{arXiv preprint arXiv:2310.18293}, 2023.

\bibitem[Chen et~al.(2022{\natexlab{b}})Chen, Zhang, Gu, Kong, Yuan, et~al.]{chen2022cross}
Zheng Chen, Yulun Zhang, Jinjin Gu, Linghe Kong, Xin Yuan, et~al.
\newblock Cross aggregation transformer for image restoration.
\newblock \emph{NeurIPS}, 2022{\natexlab{b}}.

\bibitem[Conde et~al.(2024)Conde, Geigle, and Timofte]{conde2024high}
Marcos~V Conde, Gregor Geigle, and Radu Timofte.
\newblock Instructir: High-quality image restoration following human instructions.
\newblock In \emph{ECCV}, 2024.

\bibitem[Dang et~al.(2020)Dang, Wang, Ren, and Chen]{dang2020application}
Xin Dang, Hai Wang, Jie Ren, and Le Chen.
\newblock An application performance optimization model of mobile augmented reality based on hd restoration.
\newblock In \emph{CBD}, pages 201--206. IEEE, 2020.

\bibitem[Dong et~al.(2020)Dong, Liu, Zhang, Chen, and Qiao]{dong2020fdgan}
Yu Dong, Yihao Liu, He Zhang, Shifeng Chen, and Yu Qiao.
\newblock Fd-gan: Generative adversarial networks with fusion-discriminator for single image dehazing.
\newblock In \emph{AAAI}, 2020.

\bibitem[Duan et~al.(2024)Duan, Min, Wu, Shen, and Zhai]{duan2024uniprocessor}
Huiyu Duan, Xiongkuo Min, Sijing Wu, Wei Shen, and Guangtao Zhai.
\newblock Uniprocessor: A text-induced unified low-level image processor.
\newblock In \emph{ECCV}, 2024.

\bibitem[Fan et~al.(2019)Fan, Chen, Yuan, Hua, Yu, and Chen]{fan2019dl}
Qingnan Fan, Dongdong Chen, Lu Yuan, Gang Hua, Nenghai Yu, and Baoquan Chen.
\newblock A general decoupled learning framework for parameterized image operators.
\newblock \emph{TPAMI}, 43\penalty0 (1):\penalty0 33--47, 2019.

\bibitem[Gao et~al.(2019)Gao, Tao, Shen, and Jia]{gao2019dynamic}
Hongyun Gao, Xin Tao, Xiaoyong Shen, and Jiaya Jia.
\newblock Dynamic scene deblurring with parameter selective sharing and nested skip connections.
\newblock In \emph{CVPR}, 2019.

\bibitem[Girbacia et~al.(2013)Girbacia, Butnariu, Orman, and Postelnicu]{girbacia2013virtual}
Florin Girbacia, Silviu Butnariu, Alex~Petre Orman, and Cristian~Cezar Postelnicu.
\newblock Virtual restoration of deteriorated religious heritage objects using augmented reality technologies.
\newblock \emph{European Journal of Science and Theology}, 9\penalty0 (2):\penalty0 223--231, 2013.

\bibitem[Guo et~al.(2024{\natexlab{a}})Guo, Li, Dai, Ouyang, Ren, and Xia]{guo2024mambair}
Hang Guo, Jinmin Li, Tao Dai, Zhihao Ouyang, Xudong Ren, and Shu-Tao Xia.
\newblock Mambair: A simple baseline for image restoration with state-space model.
\newblock In \emph{ECCV}, 2024{\natexlab{a}}.

\bibitem[Guo et~al.(2024{\natexlab{b}})Guo, Gao, Lu, Liu, and He]{guo2024onerestore}
Yu Guo, Yuan Gao, Yuxu Lu, Ryan~Wen Liu, and Shengfeng He.
\newblock Onerestore: A universal restoration framework for composite degradation.
\newblock In \emph{ECCV}, 2024{\natexlab{b}}.

\bibitem[Jiang et~al.(2020)Jiang, Wang, Yi, Chen, Huang, Luo, Ma, and Jiang]{jiang2020multi}
Kui Jiang, Zhongyuan Wang, Peng Yi, Chen Chen, Baojin Huang, Yimin Luo, Jiayi Ma, and Junjun Jiang.
\newblock Multi-scale progressive fusion network for single image deraining.
\newblock In \emph{CVPR}, 2020.

\bibitem[Jiang et~al.(2023)Jiang, Zhang, Xue, and Gu]{jiang2023autodir}
Yitong Jiang, Zhaoyang Zhang, Tianfan Xue, and Jinwei Gu.
\newblock Autodir: Automatic all-in-one image restoration with latent diffusion.
\newblock \emph{arXiv preprint arXiv:2310.10123}, 2023.

\bibitem[Kingma and Ba(2017)]{kingma2017adam}
Diederik~P. Kingma and Jimmy Ba.
\newblock Adam: A method for stochastic optimization.
\newblock \emph{arXiv}, 2017.

\bibitem[Kong et~al.(2023)Kong, Dong, Ge, Li, and Pan]{kong2023efficient}
Lingshun Kong, Jiangxin Dong, Jianjun Ge, Mingqiang Li, and Jinshan Pan.
\newblock Efficient frequency domain-based transformers for high-quality image deblurring.
\newblock In \emph{CVPR}, 2023.

\bibitem[Lehtinen et~al.(2018)Lehtinen, Munkberg, Hasselgren, Laine, Karras, Aittala, and Aila]{lehtinen2018noise2noise}
Jaakko Lehtinen, Jacob Munkberg, Jon Hasselgren, Samuli Laine, Tero Karras, Miika Aittala, and Timo Aila.
\newblock Noise2noise: Learning image restoration without clean data.
\newblock \emph{arXiv preprint arXiv:1803.04189}, 2018.

\bibitem[Li et~al.(2018)Li, Ren, Fu, Tao, Feng, Zeng, and Wang]{li2018benchmarking}
Boyi Li, Wenqi Ren, Dengpan Fu, Dacheng Tao, Dan Feng, Wenjun Zeng, and Zhangyang Wang.
\newblock Benchmarking single-image dehazing and beyond.
\newblock \emph{TIP}, 28\penalty0 (1):\penalty0 492--505, 2018.

\bibitem[Li et~al.(2022)Li, Liu, Hu, Wu, Lv, and Peng]{li2022airnet}
Boyun Li, Xiao Liu, Peng Hu, Zhongqin Wu, Jiancheng Lv, and Xi Peng.
\newblock {All-In-One Image Restoration for Unknown Corruption}.
\newblock In \emph{CVPR}, 2022.

\bibitem[Li et~al.(2023)Li, Lei, Ma, Zhang, and Shan]{li2023pip}
Zilong Li, Yiming Lei, Chenglong Ma, Junping Zhang, and Hongming Shan.
\newblock Prompt-in-prompt learning for universal image restoration.
\newblock \emph{arXiv preprint arXiv:2312.05038}, 2023.

\bibitem[Liang et~al.(2021)Liang, Cao, Sun, Zhang, Van~Gool, and Timofte]{liang2021swinir}
Jingyun Liang, Jiezhang Cao, Guolei Sun, Kai Zhang, Luc Van~Gool, and Radu Timofte.
\newblock Swinir: Image restoration using swin transformer.
\newblock In \emph{ICCV}, 2021.

\bibitem[Liu et~al.(2022)Liu, Xie, Zhang, Yuan, Chen, Zhou, Li, and Tian]{liu2022tape}
Lin Liu, Lingxi Xie, Xiaopeng Zhang, Shanxin Yuan, Xiangyu Chen, Wengang Zhou, Houqiang Li, and Qi Tian.
\newblock Tape: Task-agnostic prior embedding for image restoration.
\newblock In \emph{ECCV}, 2022.

\bibitem[Liu et~al.(2021)Liu, Lin, Cao, Hu, Wei, Zhang, Lin, and Guo]{liu2021swin}
Ze Liu, Yutong Lin, Yue Cao, Han Hu, Yixuan Wei, Zheng Zhang, Stephen Lin, and Baining Guo.
\newblock Swin transformer: Hierarchical vision transformer using shifted windows.
\newblock In \emph{ICCV}, 2021.

\bibitem[Luo et~al.(2023)Luo, Zhao, Wei, Chen, Lu, Xie, Wang, Xiong, Lu, and Zhang]{luo2023wm}
Yulin Luo, Rui Zhao, Xiaobao Wei, Jinwei Chen, Yijie Lu, Shenghao Xie, Tianyu Wang, Ruiqin Xiong, Ming Lu, and Shanghang Zhang.
\newblock Wm-moe: Weather-aware multi-scale mixture-of-experts for blind adverse weather removal.
\newblock \emph{arXiv preprint arXiv:2303.13739}, 2023.

\bibitem[Luo et~al.(2024)Luo, Gustafsson, Zhao, Sj{\"o}lund, and Sch{\"o}n]{luo2023daclip}
Ziwei Luo, Fredrik~K Gustafsson, Zheng Zhao, Jens Sj{\"o}lund, and Thomas~B Sch{\"o}n.
\newblock Controlling vision-language models for multi-task image restoration.
\newblock In \emph{ICLR}, 2024.

\bibitem[Ma et~al.(2016)Ma, Duanmu, Wu, Wang, Yong, Li, and Zhang]{ma2016waterloo}
Kede Ma, Zhengfang Duanmu, Qingbo Wu, Zhou Wang, Hongwei Yong, Hongliang Li, and Lei Zhang.
\newblock Waterloo exploration database: New challenges for image quality assessment models.
\newblock \emph{TIP}, 26\penalty0 (2):\penalty0 1004--1016, 2016.

\bibitem[Martin et~al.(2001)Martin, Fowlkes, Tal, and Malik]{martin2001database}
David Martin, Charless Fowlkes, Doron Tal, and Jitendra Malik.
\newblock A database of human segmented natural images and its application to evaluating segmentation algorithms and measuring ecological statistics.
\newblock In \emph{ICCV}, 2001.

\bibitem[Mou et~al.(2022)Mou, Wang, and Zhang]{mou2022dgunet}
Chong Mou, Qian Wang, and Jian Zhang.
\newblock Deep generalized unfolding networks for image restoration.
\newblock In \emph{CVPR}, 2022.

\bibitem[Nah et~al.(2017)Nah, Hyun~Kim, and Mu~Lee]{nah2017deep}
Seungjun Nah, Tae Hyun~Kim, and Kyoung Mu~Lee.
\newblock Deep multi-scale convolutional neural network for dynamic scene deblurring.
\newblock In \emph{CVPR}, 2017.

\bibitem[Pan et~al.(2016)Pan, Sun, Pfister, and Yang]{pan2016blind}
Jinshan Pan, Deqing Sun, Hans Pfister, and Ming-Hsuan Yang.
\newblock Blind image deblurring using dark channel prior.
\newblock In \emph{CVPR}, 2016.

\bibitem[Potlapalli et~al.(2023)Potlapalli, Zamir, Khan, and Khan]{potlapalli2023promptir}
Vaishnav Potlapalli, Syed~Waqas Zamir, Salman Khan, and Fahad Khan.
\newblock Promptir: Prompting for all-in-one image restoration.
\newblock In \emph{NeurIPS}, 2023.

\bibitem[Puigcerver et~al.(2024)Puigcerver, Ruiz, Mustafa, and Houlsby]{puigcerver2024softmoe}
Joan Puigcerver, Carlos~Riquelme Ruiz, Basil Mustafa, and Neil Houlsby.
\newblock From sparse to soft mixtures of experts.
\newblock In \emph{ICLR}, 2024.

\bibitem[Qu et~al.(2019)Qu, Chen, Huang, and Xie]{qu2019enhanced}
Yanyun Qu, Yizi Chen, Jingying Huang, and Yuan Xie.
\newblock Enhanced pix2pix dehazing network.
\newblock In \emph{CVPR}, 2019.

\bibitem[Raposo et~al.(2024)Raposo, Ritter, Richards, Lillicrap, Humphreys, and Santoro]{raposo2024mod}
David Raposo, Sam Ritter, Blake Richards, Timothy Lillicrap, Peter~Conway Humphreys, and Adam Santoro.
\newblock Mixture-of-depths: Dynamically allocating compute in transformer-based language models, 2024.

\bibitem[Ren et~al.(2019)Ren, Zuo, Hu, Zhu, and Meng]{ren2019progressive}
Dongwei Ren, Wangmeng Zuo, Qinghua Hu, Pengfei Zhu, and Deyu Meng.
\newblock Progressive image deraining networks: A better and simpler baseline.
\newblock In \emph{CVPR}, 2019.

\bibitem[Ren et~al.(2018)Ren, Ma, Zhang, Pan, Cao, Liu, and Yang]{ren2018gated}
Wenqi Ren, Lin Ma, Jiawei Zhang, Jinshan Pan, Xiaochun Cao, Wei Liu, and Ming-Hsuan Yang.
\newblock Gated fusion network for single image dehazing.
\newblock In \emph{CVPR}, 2018.

\bibitem[Ren et~al.(2020)Ren, Pan, Zhang, Cao, and Yang]{ren2020singledehazing}
Wenqi Ren, Jinshan Pan, Hua Zhang, Xiaochun Cao, and Ming-Hsuan Yang.
\newblock Single image dehazing via multi-scale convolutional neural networks with holistic edges.
\newblock In \emph{ICCV}, 2020.

\bibitem[Riquelme et~al.(2021)Riquelme, Puigcerver, Mustafa, Neumann, Jenatton, Susano~Pinto, Keysers, and Houlsby]{riquelme2021scaling}
Carlos Riquelme, Joan Puigcerver, Basil Mustafa, Maxim Neumann, Rodolphe Jenatton, Andr{\'e} Susano~Pinto, Daniel Keysers, and Neil Houlsby.
\newblock Scaling vision with sparse mixture of experts.
\newblock \emph{NeurIPS}, 2021.

\bibitem[Ronneberger et~al.(2015)Ronneberger, Fischer, and Brox]{ronneberger2015unet}
Olaf Ronneberger, Philipp Fischer, and Thomas Brox.
\newblock {U-Net}: Convolutional networks for biomedical image segmentation.
\newblock In \emph{MICCAI}, pages 234--241. Springer, 2015.

\bibitem[Saggio et~al.(2011)Saggio, Borra, et~al.]{saggio2011augmented}
Giovanni Saggio, Davide Borra, et~al.
\newblock Augmented reality for restoration/reconstruction of artefacts with artistic or historical value.
\newblock In \emph{Augmented reality: some emerging application areas}, pages 59--86. InTech Publications, 2011.

\bibitem[Shazeer et~al.(2017)Shazeer, Mirhoseini, Maziarz, Davis, Le, Hinton, and Dean]{shazeer2017outrageously}
Noam Shazeer, Azalia Mirhoseini, Krzysztof Maziarz, Andy Davis, Quoc Le, Geoffrey Hinton, and Jeff Dean.
\newblock Outrageously large neural networks: The sparsely-gated mixture-of-experts layer, 2017.

\bibitem[Tai et~al.(2017)Tai, Yang, Liu, and Xu]{tai2017memnet}
Ying Tai, Jian Yang, Xiaoming Liu, and Chunyan Xu.
\newblock Memnet: A persistent memory network for image restoration.
\newblock In \emph{ICCV}, 2017.

\bibitem[Tian et~al.(2020)Tian, Xu, and Zuo]{tian2000brdnet}
Chunwei Tian, Yong Xu, and Wangmeng Zuo.
\newblock Image denoising using deep cnn with batch renormalization.
\newblock \emph{NN}, 2020.

\bibitem[Valanarasu et~al.(2022)Valanarasu, Yasarla, and Patel]{valanarasu2022transweather}
Jeya Maria~Jose Valanarasu, Rajeev Yasarla, and Vishal~M Patel.
\newblock Transweather: Transformer-based restoration of images degraded by adverse weather conditions.
\newblock In \emph{CVPR}, 2022.

\bibitem[Wang et~al.(2023)Wang, Pan, Wang, Dong, Wang, Ju, and Chen]{wang2023promptrestorer}
Cong Wang, Jinshan Pan, Wei Wang, Jiangxin Dong, Mengzhu Wang, Yakun Ju, and Junyang Chen.
\newblock Promptrestorer: A prompting image restoration method with degradation perception.
\newblock In \emph{NeurIPS}, 2023.

\bibitem[Wang et~al.(2024)Wang, Zhang, Shao, Luo, Stenger, Lu, Kim, Liu, and Li]{wang2024gridformer}
Tao Wang, Kaihao Zhang, Ziqian Shao, Wenhan Luo, Bjorn Stenger, Tong Lu, Tae-Kyun Kim, Wei Liu, and Hongdong Li.
\newblock Gridformer: Residual dense transformer with grid structure for image restoration in adverse weather conditions.
\newblock \emph{ICCV}, 2024.

\bibitem[Wang et~al.(2022)Wang, Cun, Bao, Zhou, Liu, and Li]{wang2022uformer}
Zhendong Wang, Xiaodong Cun, Jianmin Bao, Wengang Zhou, Jianzhuang Liu, and Houqiang Li.
\newblock Uformer: A general u-shaped transformer for image restoration.
\newblock In \emph{CVPR}, 2022.

\bibitem[Wei et~al.(2018)Wei, Wang, Yang, and Liu]{wei2018deep}
Chen Wei, Wenjing Wang, Wenhan Yang, and Jiaying Liu.
\newblock Deep retinex decomposition for low-light enhancement.
\newblock \emph{arXiv preprint arXiv:1808.04560}, 2018.

\bibitem[Wu et~al.(2024)Wu, Jiang, Jiang, and Liu]{wu2024art}
Gang Wu, Junjun Jiang, Kui Jiang, and Xianming Liu.
\newblock Harmony in diversity: Improving all-in-one image restoration via multi-task collaboration.
\newblock In \emph{ACMMM}, 2024.

\bibitem[Wu et~al.(2021)Wu, Qu, Lin, Zhou, Qiao, Zhang, Xie, and Ma]{wu2021contrastivedehazing}
Haiyan Wu, Yanyun Qu, Shaohui Lin, Jian Zhou, Ruizhi Qiao, Zhizhong Zhang, Yuan Xie, and Lizhuang Ma.
\newblock Contrastive learning for compact single image dehazing.
\newblock In \emph{CVPR}, 2021.

\bibitem[Yang et~al.(2020)Yang, Yang, Fu, Lu, and Guo]{yang2020learning}
Fuzhi Yang, Huan Yang, Jianlong Fu, Hongtao Lu, and Baining Guo.
\newblock Learning texture transformer network for image super-resolution.
\newblock In \emph{CVPR}, 2020.

\bibitem[Yang et~al.(2024{\natexlab{a}})Yang, Pan, Yang, and Liang]{yang2024ldr}
Hao Yang, Liyuan Pan, Yan Yang, and Wei Liang.
\newblock Language-driven all-in-one adverse weather removal.
\newblock In \emph{CVPR}, 2024{\natexlab{a}}.

\bibitem[Yang et~al.(2024{\natexlab{b}})Yang, Chen, Qian, Yi, Zhang, Zhao, Wei, and Xu]{yang2024amir}
Zhiwen Yang, Haowei Chen, Ziniu Qian, Yang Yi, Hui Zhang, Dan Zhao, Bingzheng Wei, and Yan Xu.
\newblock All-in-one medical image restoration via task-adaptive routing.
\newblock In \emph{MICCAI}, 2024{\natexlab{b}}.

\bibitem[Yasarla and Patel(2019)]{yasarla2019uncertainty}
Rajeev Yasarla and Vishal~M Patel.
\newblock Uncertainty guided multi-scale residual learning-using a cycle spinning cnn for single image de-raining.
\newblock In \emph{CVPR}, 2019.

\bibitem[Yu et~al.(2022)Yu, Wang, Dong, Tang, and Loy]{yu2022:pathrestore}
Ke Yu, Xintao Wang, Chao Dong, Xiaoou Tang, and Chen~Change Loy.
\newblock Path-restore: Learning network path selection for image restoration.
\newblock \emph{TPAMI}, 2022.

\bibitem[Yu et~al.(2024)Yu, Zhou, Li, and Zhu]{yu2024multi}
Xiaoyan Yu, Shen Zhou, Huafeng Li, and Liehuang Zhu.
\newblock Multi-expert adaptive selection: Task-balancing for all-in-one image restoration.
\newblock \emph{arXiv preprint arXiv:2407.19139}, 2024.

\bibitem[Zamfir et~al.(2024{\natexlab{a}})Zamfir, Wu, Mehta, Paudel, Zhang, and Timofte]{zamfir2024efficient}
Eduard Zamfir, Zongwei Wu, Nancy Mehta, Danda~Pani Paudel, Yulun Zhang, and Radu Timofte.
\newblock Efficient degradation-aware any image restoration.
\newblock \emph{arXiv preprint arXiv:2405.15475}, 2024{\natexlab{a}}.

\bibitem[Zamfir et~al.(2024{\natexlab{b}})Zamfir, Wu, Mehta, Zhang, and Timofte]{zamfir2024details}
Eduard Zamfir, Zongwei Wu, Nancy Mehta, Yulun Zhang, and Radu Timofte.
\newblock See more details: Efficient image super-resolution by experts mining.
\newblock In \emph{ICML}. PMLR, 2024{\natexlab{b}}.

\bibitem[Zamir et~al.(2021)Zamir, Arora, Khan, Hayat, Khan, Yang, and Shao]{zamir2021pmrnet}
Syed~Waqas Zamir, Aditya Arora, Salman Khan, Munawar Hayat, Fahad~Shahbaz Khan, Ming-Hsuan Yang, and Ling Shao.
\newblock Multi-stage progressive image restoration.
\newblock In \emph{CVPR}, 2021.

\bibitem[Zamir et~al.(2022)Zamir, Arora, Khan, Hayat, Khan, and Yang]{Zamir2021Restormer}
Syed~Waqas Zamir, Aditya Arora, Salman Khan, Munawar Hayat, Fahad~Shahbaz Khan, and Ming-Hsuan Yang.
\newblock Restormer: Efficient transformer for high-resolution image restoration.
\newblock In \emph{CVPR}, 2022.

\bibitem[Zhang et~al.(2023)Zhang, Huang, Yao, Yang, Yu, Zhou, and Zhao]{zhang2023ingredient}
Jinghao Zhang, Jie Huang, Mingde Yao, Zizheng Yang, Hu Yu, Man Zhou, and Feng Zhao.
\newblock Ingredient-oriented multi-degradation learning for image restoration.
\newblock In \emph{CVPR}, 2023.

\bibitem[Zhang et~al.(2017)Zhang, Zuo, Gu, and Zhang]{zhang2017learning}
Kai Zhang, Wangmeng Zuo, Shuhang Gu, and Lei Zhang.
\newblock Learning deep cnn denoiser prior for image restoration.
\newblock In \emph{CVPR}, 2017.

\bibitem[Zhang et~al.(2019)Zhang, Li, Li, Zhong, and Fu]{zhang2019residual}
Yulun Zhang, Kunpeng Li, Kai Li, Bineng Zhong, and Yun Fu.
\newblock Residual non-local attention networks for image restoration.
\newblock \emph{arXiv preprint arXiv:1903.10082}, 2019.

\bibitem[Zhao et~al.(2023)Zhao, Gou, Li, Peng, Lv, and Peng]{zhao2023comprehensive}
Haiyu Zhao, Yuanbiao Gou, Boyun Li, Dezhong Peng, Jiancheng Lv, and Xi Peng.
\newblock Comprehensive and delicate: An efficient transformer for image restoration.
\newblock In \emph{CVPR}, 2023.

\bibitem[Zhou et~al.(2022)Zhou, Lei, Liu, Du, Huang, Zhao, Dai, Le, Laudon, et~al.]{zhou2022mixture}
Yanqi Zhou, Tao Lei, Hanxiao Liu, Nan Du, Yanping Huang, Vincent Zhao, Andrew~M Dai, Quoc~V Le, James Laudon, et~al.
\newblock Mixture-of-experts with expert choice routing.
\newblock \emph{NeurIPS}, 2022.

\bibitem[Zhu et~al.(2023)Zhu, Wang, Fu, Yang, Guo, Dai, Qiao, and Hu]{zhu2023Weather}
Yurui Zhu, Tianyu Wang, Xueyang Fu, Xuanyu Yang, Xin Guo, Jifeng Dai, Yu Qiao, and Xiaowei Hu.
\newblock Learning weather-general and weather-specific features for image restoration under multiple adverse weather conditions.
\newblock In \emph{CVPR}, 2023.

\bibitem[Özdenizci and Legenstein(2023)]{oezdenizci2022WeatherDiff}
Ozan Özdenizci and Robert Legenstein.
\newblock Restoring vision in adverse weather conditions with patch-based denoising diffusion models.
\newblock \emph{TPAMI}, 2023.

\end{thebibliography}
}
\end{document}